\documentclass{article}

\usepackage{iclr2027_conference,times}
\usepackage[T1]{fontenc}
\usepackage{amsmath,amssymb,amsthm}
\usepackage{microtype}
\usepackage{booktabs}
\usepackage{multirow}
\usepackage{float}
\usepackage{graphicx}
\usepackage{subcaption}
\usepackage{xcolor}
\usepackage[hidelinks]{hyperref}
\usepackage{url}
\usepackage{listings}
\title{Knowing Is Not Choosing:\\
What Explicit Verification Adds\\
Beyond Generative Preference}

\author{Yilong Li, Chengpo Yan, Aayan Arish, Suman Banerjee \\
Department of Computer Science\\
University of Wisconsin-Madison\\
Madison, WI 53703, USA \\
\texttt{\{yilong,chengpo,suman\}@cs.wisc.edu} \\
}

\iclrfinalcopy

\begin{document}

\maketitle

\begin{abstract}
Generating a correct answer does not mean that a language model will select it. We separate factual recall into three steps: generating a correct candidate, ranking the available candidates, and selecting the final answer. Pre-generation readouts predict factual recall and which questions sampling will cover across three model families, but say little about whether an available correct answer will ultimately be selected. Explicit verification with $P(\mathrm{True})$ improves within-question ranking over mean log-likelihood in Gemma, Qwen3, and Llama, with AUROC gains of $0.08$--$0.12$. In a prospectively defined Gemma cohort, verification raises plurality accuracy by about $5$ points, and still gains about $2$ points over chat-template likelihood, a stronger generative baseline. The advantage is strongest for relations with common-answer priors and depends on access to the entity; masking the entity removes the ranking advantage in larger Qwen models. Finally, the measured benefit depends on how correctness is defined: recall-oriented reference matching can credit option lists favored by likelihood and substantially understate the improvement seen under human semantic judgments.
Prior work shows that models can carry latent factual knowledge and judge candidate answers; we show that these capabilities do not collapse into a single notion of ``knowing,'' and trace where information is gained, lost, or mismeasured between availability, ranking, and final choice.
\end{abstract}

\vspace{-0.6ex}
\section{Introduction}
\label{sec:intro}
\vspace{-0.6ex}

Repeated sampling can expose correct answers that a language model does not return on its first attempt. Prior work shows that candidate coverage can grow rapidly with inference-time sampling, especially when successful outputs can be externally verified \citep{brown2024monkeys}. Factual question answering is harder: the model may generate a correct candidate yet still select an incorrect alternative. We ask \emph{when factual competence translates into a correct final answer}.
Sampling increases coverage---the fraction of questions with a correct candidate---while explicit verification changes which candidate is selected. We study when verification converts this coverage into correct answers and how much of its gain remains against stronger generative baselines and under different correctness criteria. To do so, we separate expected recall before generation, within-question candidate ranking, and final selection.

Across Gemma, Qwen3, and Llama, pre-generation readouts improve held-out prediction of expected factual recall beyond measured exposure and identity features: on $3{,}840$ entities per model, held-out log loss falls by $3.5\%$, $1.7\%$, and $3.9\%$ relative to the baseline (Section~\ref{sec:construct}). Without refitting, the same readouts also predict whether up to $16$ sampled candidates contain a correct answer, with AUROC $0.7815$ in Gemma, $0.8165$ in Llama, and $0.6898$ in Qwen3. Each readout is computed before generation, and it adds little about whether an available correct candidate is ultimately selected: in Gemma and Llama, adding the readout to exposure, identity features, and a common-answer prior raises AUROC by $0.040$ and $0.041$ for coverage, but only $0.006$--$0.015$ for plurality success (Section~\ref{sec:know-choose}).

\begin{figure}[htb]
\centering
\includegraphics[width=0.8\linewidth,pagebox=cropbox]{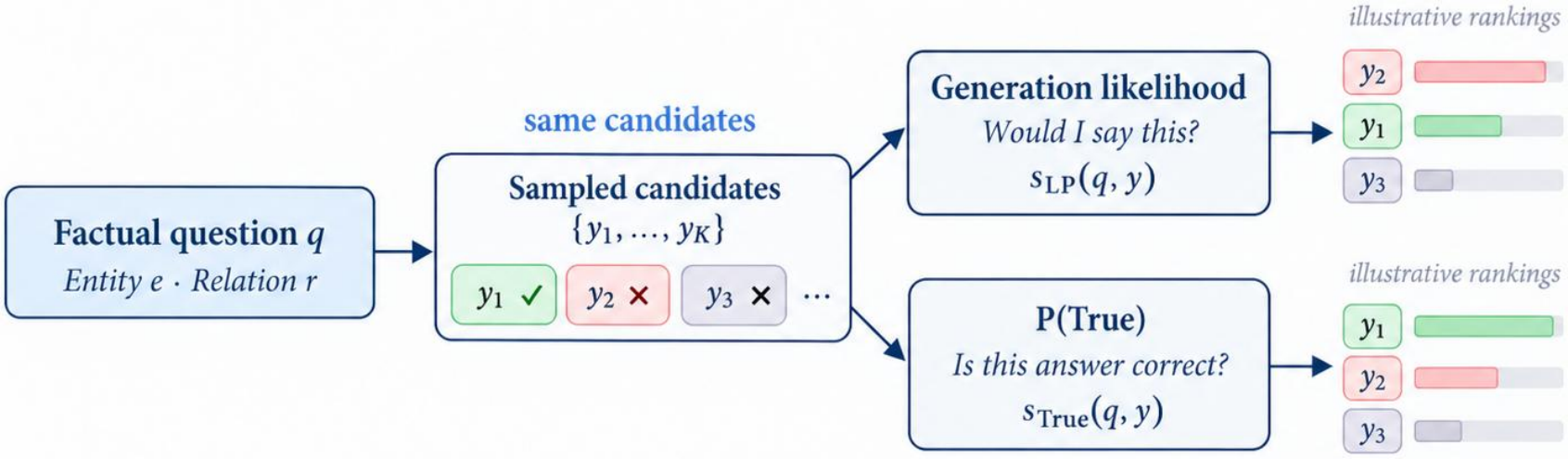}
\caption{\textbf{Same candidates, different rankings.} Generation likelihood
reflects \emph{``Would I say this?''}; $P(\mathrm{True})$ asks the same model
whether the answer is correct (prompt in Appendix~\ref{app:prompts}) and takes
its \emph{Yes} probability, normalized over \emph{Yes} and \emph{No}. Box texts
paraphrase the two scores; rankings are illustrative.}
\label{fig:main}
\vspace{-1ex}
\end{figure}

\begin{figure}[htb]
\centering
\captionsetup[subfigure]{font=small,labelfont=bf,justification=raggedright,
singlelinecheck=false,skip=4pt}
\begin{subfigure}[t]{0.345\linewidth}
\centering
\includegraphics[width=\linewidth]{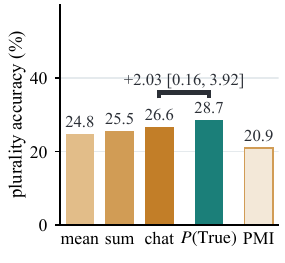}
\caption{\textbf{Generative rescoring explains part, but not all, of the gap.}}
\end{subfigure}\hfill
\begin{subfigure}[t]{0.34\linewidth}
\centering
\includegraphics[width=\linewidth]{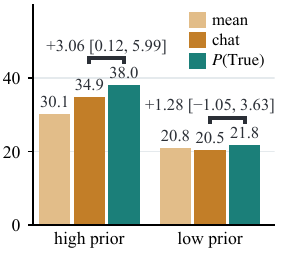}
\caption{\textbf{The gain over chat-template likelihood is larger in high-prior relations.}}
\end{subfigure}\hfill
\begin{subfigure}[t]{0.28\linewidth}
\centering
\includegraphics[width=\linewidth]{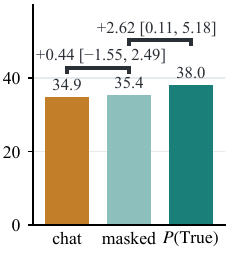}
\caption{\textbf{Entity masking removes most of the high-prior gain.}}
\end{subfigure}
\caption{\textbf{What verification adds beyond generative baselines.}
Gemma plurality accuracy on 541 questions under semantic labels.
(a) Generative baselines versus $P(\mathrm{True})$.
(b) Results split by common-answer prior.
(c) Entity masking on high-prior questions.
Brackets show paired differences with 95\% intervals.}
\label{fig:decomposition}
\vspace{-1ex}
\end{figure}

Choosing among candidates requires a score for each answer. Likelihood reflects the model's generative preference, whereas explicit verification asks whether a candidate is correct (Figure~\ref{fig:main}). Consistent with \citet{gekhman2025insideout}, $P(\mathrm{True})$ ranks answers to the same question better than likelihood, improving pair-weighted AUROC by $0.102$ in Gemma, $0.119$ in Llama, and $0.078$ in Qwen3. In the Gemma selection cohort, increasing the candidate budget from $1$ to $16$ raises coverage from $17.4\%$ to $37.7\%$, yet likelihood-based plurality reaches only $23.1\%$ accuracy. Verification raises this to $28.1\%$, a $4.99$-point gain over mean log-likelihood ($95\%$ CI $[2.7,7.3]$), and increases conversion from $61.3\%$ to $74.5\%$. The gain persists across prompt variants, confidence-weighted voting, fresh samples, and pools filtered to remove option lists and other obvious non-answers. Across the other models in Table~\ref{tab:models}, verification ranks better in all but Gemma~2 27B-IT and Qwen3.5-2B, while its plurality gain ranges from $-1.61$ points in Qwen3.5-2B to $3.40$ in Qwen3.8-27B.

\begin{table}[htb]
\centering
\footnotesize
\setlength{\tabcolsep}{4pt}
\caption{\textbf{Verification improves ranking more consistently than final selection.}
Stricter criterion, $K=16$; accuracy runs from mean log-likelihood to $P(\mathrm{True})$,
and $\Delta$AUROC (pair-weighted) and gain are their differences with paired $95\%$
intervals. Gemma 2B is the prospective cohort; the other rows reuse its questions, and
the four newer Qwen models are post hoc (Appendix~\ref{app:provenance}). Coverage and
conversion: Appendices~\ref{app:selection-tables}, \ref{app:llama-selection},
and~\ref{app:qwen-selection}.}
\label{tab:models}
\begin{tabular}{lrr@{\ }lcr@{\ }l}
\toprule
Model & $N$ & \multicolumn{2}{c}{$\Delta$AUROC} & Plurality accuracy (\%) & \multicolumn{2}{c}{Gain (points)} \\
\midrule
Gemma 2B  & $541$ & $+0.102$ & $[0.056,0.149]$ & $23.11{\to}28.10$ & $+4.99$ & $[2.7,7.3]$ \\
Gemma 9B  & $581$ & $+0.032$ & $[0.006,0.058]$ & $43.55{\to}43.72$ & $+0.17$ & $[-0.86,1.20]$ \\
Gemma 27B & $564$ & $+0.030$ & $[-0.033,0.091]$ & $48.05{\to}48.67$ & $+0.62$ & $[-1.03,2.28]$ \\
Llama 8B  & $598$ & $+0.119$ & $[0.084,0.154]$ & $43.48{\to}45.15$ & $+1.67$ & $[0.05,3.30]$ \\
Qwen3.5 2B  & $528$ & $+0.008$ & $[-0.053,0.069]$ & $11.55{\to}9.94$ & $-1.61$ & $[-3.29,0.00]$ \\
Qwen3.5 4B  & $565$ & $+0.139$ & $[0.101,0.176]$ & $19.47{\to}21.59$ & $+2.12$ & $[0.00,4.26]$ \\
Qwen3 8B    & $600$ & $+0.078$ & $[0.038,0.118]$ & $26.17{\to}27.33$ & $+1.17$ & $[-0.75,3.10]$ \\
Qwen3.6 27B & $599$ & $+0.180$ & $[0.149,0.213]$ & $30.05{\to}33.39$ & $+3.34$ & $[0.92,5.82]$ \\
Qwen3.8 27B & $594$ & $+0.149$ & $[0.111,0.186]$ & $24.75{\to}28.14$ & $+3.40$ & $[0.97,5.89]$ \\
\bottomrule
\end{tabular}
\vspace{-1ex}
\end{table}

Part of Gemma's gain reflects how likelihood is scored (Figure~\ref{fig:decomposition}). Chat-template likelihood scores each candidate as the model's reply to an instruction to return only the requested answer; under semantic labels, it raises plurality accuracy from $24.77\%$ to $26.62\%$, while verification reaches $28.65\%$, remaining $2.03$ points higher ($[0.16,3.92]$). This residual is concentrated in the four high-prior relations ($3.06$ points versus $1.28$ in the other six) and depends on the entity: replacing the entity with its type reduces verification's advantage over chat-template likelihood to $0.44$ points. The stricter criterion yields the same pattern with slightly larger gaps, and two 27B Qwen models likewise lose verification's ranking advantage when the entity is masked (Section~\ref{sec:rescoring}; Appendix~\ref{app:rescoring}). The observed selection gain also depends on the correctness criterion: on identical outputs, verification's plurality gain over mean log-likelihood is $1.29$ points under reference matching but $4.99$ under the stricter criterion. Semantic and human labels place the gain at $3.88$ ($[1.6,6.1]$) and $3.70$ points ($[1.31,6.08]$), respectively, so most of the effect is lost under reference matching (Section~\ref{sec:selection-label-sensitivity}). The main source is option lists containing the reference answer, which likelihood selects much more often; $18$ of the $19$ reference-matching errors on likelihood's argmax outputs are such lists.

We disentangle factual competence into candidate availability, within-question ranking, and final selection, showing that knowing which questions are likely answerable is not enough to identify or select the correct answer.

\vspace{-0.6ex}
\section{Related Work}
\label{sec:related}
\vspace{-0.6ex}

\noindent\textbf{Repeated sampling and answer selection.}
Repeated sampling increases candidate coverage, but imperfect
selection limits how much of this gain reaches the final answer
\citep{brown2024monkeys,stroebl2026limits}; coverage gains also depend on the
baseline used for comparison \citep{yona2025keep}. Self-consistency selects by
plurality \citep{wang2023selfconsistency}, while
\citet{taubenfeld2025confidence} use confidence-weighted voting.
\citet{gekhman2025insideout} compare generation probability, length-normalized
probability, and $P(\mathrm{True})$ for ranking sampled answers, finding that
better pairwise ranking need not yield better final selection. We study this
gap on fixed candidate pools, including stronger generative baselines and the
effect of the correctness criterion.

\vspace{-0.5ex}
\noindent\textbf{Knowledge awareness.}
\citet{kadavath2022know} distinguish $P(\mathrm{IK})$, which estimates whether a
model can answer a question, from $P(\mathrm{True})$, which evaluates a specific
answer. \citet{ferrando2025entity} identify an entity-recognition signal associated with
factual recall and refusal, \citet{gottesman2024estimating} estimate how much a
model knows about an entity from its subject representations before generation,
and \citet{kumaran2026causal} provide causal evidence that models use confidence
to drive behavior. Factual accuracy also tracks exposure and entity popularity
\citep{kandpal2023longtail,mallen2023popqa}. We test whether such pre-generation
readouts predict factual recall beyond measured exposure and identity features,
which questions sampling will cover, and which covered questions end with a
correct final answer.

\vspace{-0.5ex}
\noindent\textbf{Answer-level correctness signals.}
Hidden states encode signals related to factual correctness
\citep{marks2023geometry,orgad2025know,obeso2025realtime}, although a signal of
recall need not imply that a generated answer is true
\citep{cheang2026really}. Other approaches estimate uncertainty from variation
across sampled responses
\citep{manakul2023selfcheckgpt,kuhn2023semantic,farquhar2024semantic,kossen2024sep},
or score a candidate directly through model-based verification
\citep{li2024benchmarking,zhang2025genrm}. Our focus is the latter setting:
given the same generated candidates, we compare generative preference with
explicit factual verification and measure how each affects final selection.

\vspace{-0.5ex}
\noindent\textbf{Correctness criteria and benchmark artifacts.}
Evaluation itself can distort measured performance: benchmark construction may
leak reference information, and lexical matching can misjudge free-form answers
\citep{hussain2026parallax,kamalloo2023evaluating}. We examine a complementary
effect in scoring selected outputs. On identical selected answers, recall-oriented
reference matching reduces the measured verification gain;
semantic and human labels trace the difference mostly to unresolved option
lists that reference matching credits.
\vspace{-0.5ex}
\vspace{-0.6ex}
\section{Methodology}
\label{sec:setup}
\vspace{-0.6ex}

\vspace{-0.6ex}
\subsection{Measuring factual competence}
\vspace{-0.6ex}
A factual query $q=(e,r)$ asks for relation $r$ of entity $e$. We separate
three observable stages of inference: whether a correct candidate is generated,
whether a score can identify it among alternatives, and whether it is ultimately
selected. We treat $\mathcal Y_{q,K}$ as a multiset of sampled candidates,
retaining duplicates for voting; the selector returns one of its members. Let
$\ell(q,y)\in\{0,1\}$ denote correctness under the specified criterion. We define
\[
H_{q,K}=\max_{y\in\mathcal Y_{q,K}}\ell(q,y),\qquad
C_K=\mathbb E[H_{q,K}],
\]
where $C_K$ is candidate coverage, and
\[
A^\pi_K=
\mathbb E\!\left[\ell\!\left(q,\pi(\mathcal Y_{q,K})\right)\right]
\]
as the final accuracy of selector $\pi$. Their ratio, the conversion
\[
R^\pi_K=\frac{A^\pi_K}{C_K},
\]
measures how often a selector returns a correct answer when the pool contains one.
Equivalently,
\begin{equation}
1-A^\pi_K=(1-C_K)+(C_K-A^\pi_K),
\label{eq:selection-gap}
\end{equation}
separating missing coverage from selection error.

To measure ranking independently of final selection, we compare correct and
incorrect answers only within the same question. For a score $S$,
\[
a_q(S)=
\frac{1}{n_q^+n_q^-}
\sum_{y^+}\sum_{y^-}
\phi\!\left(S(q,y^+),S(q,y^-)\right),
\]
where $\phi$ assigns $1$ to a correctly ordered pair, $1/2$ to a tie, and $0$
otherwise. We compute $a_q$ only for mixed questions, with $n_q^+$ correct and
$n_q^-$ incorrect candidates. We aggregate equally across questions
(query-macro) or with weights $n_q^+n_q^-$ (pair-weighted AUROC)
\citep{taubenfeld2025confidence}.

\vspace{-0.6ex}
\subsection{Models and cohorts}
\vspace{-0.6ex}
We study Gemma~2 2B-IT \citep{gemmateam2024gemma2}, Qwen3-8B \citep{yang2025qwen3}, and Llama~3.1 8B-Instruct \citep{grattafiori2024llama3}, hereafter Gemma, Qwen3, and Llama. Recall prediction uses fixed pre-generation readouts: a layer-$16$ class-mean direction for Gemma, a separately fitted layer-$15$ direction for Llama, and negatively oriented Qwen-Scope feature $58682$ at layer $23$ for Qwen3 \citep{deng2026qwenscope}. Gemma and Llama are read at the final entity token of an entity-only prompt, whereas Qwen3 is read at the final entity mention in the factual query; the Qwen-Scope encoder is transferred unchanged from the Qwen3-8B base model. Each recall study uses $3{,}840$ Wikidata entities \citep{vrandecic2014wikidata} spanning the four types of \citet{ferrando2025entity}, split into fit, calibration, and confirmation sets. The prospective selection study uses $600$ fresh factual questions over $489$ entities, with at most two questions per entity and no development-set overlap, covering nine name-valued relations and song publication year. We designate country, teams, genres, and publication year as \emph{high-prior} because their $16$ most common answers cover at least $10\%$ of questions; the remaining six are \emph{low-prior}. For Gemma, we sample $16$ responses per question at $T=1$, top-$k=50$, no nucleus truncation, and a $24$-token limit; a pre-specified format filter yields $6{,}864$ candidates from $541$ eligible questions. Llama and Qwen3 use the same questions at $T=1$ without top-$k$ or nucleus truncation, yielding $598$ and $600$ eligible questions. We additionally evaluate Gemma~2 9B-IT/27B-IT and Qwen3.5-2B, Qwen3.5-4B, Qwen3.6-27B, and Qwen3.8-27B on the same selection cohort \citep{gemmateam2024gemma2,qwen35,qwen36,qwen38}. Full readout, decoding, and cohort details are in Appendices~\ref{app:reproduction}, \ref{app:qwen}, \ref{app:llama}, \ref{app:scoring}, \ref{app:gemma27b}, \ref{app:qwen-family}, and~\ref{app:cohorts}.

\vspace{-0.6ex}
\subsection{Candidate scores and selection}
\vspace{-0.6ex}

Our primary likelihood score is mean teacher-forced log probability of the
decoded answer (mean log-likelihood, abbreviated likelihood),
\[
s_{\rm LP}(q,y)=
\frac{1}{|y|}
\sum_t \log p(y_t\mid q,y_{<t}),
\]
excluding prompt and padding tokens. Summed log-likelihood omits the length
normalization.

For explicit verification, the same model receives the question and one
candidate answer and is asked:
\emph{Is the proposed answer factually correct? Reply Yes or No.}
(Appendix~\ref{app:prompts} gives every prompt in full.)
We define
\[
s_{\rm True}(q,y)=
\frac{\exp z_{\rm Yes}}
     {\exp z_{\rm Yes}+\exp z_{\rm No}},
\]
using the first-response logits of the single \emph{Yes} and \emph{No} tokens
(Appendix~\ref{app:prompts}, Table~\ref{tab:label-tokens}). No reference answer is provided.
We use $P(\mathrm{True})$ as shorthand for this ranking score.

We evaluate both argmax selection, which returns the highest-scoring candidate,
and plurality voting \citep{wang2023selfconsistency}. After lowercasing,
deleting every character other than ASCII letters, digits, and spaces, and
truncating to $60$ characters, we retain the most frequent
answer groups and return the highest-scoring original candidate among them.
Thus a group's tie-break score is its maximum candidate score. Remaining
score ties are averaged uniformly. The candidate pools are identical across evaluators. Prompt variants
and confidence-weighted voting are reported in
Appendices~\ref{app:prompt-robustness} and~\ref{app:weighted-voting}.

\vspace{-0.6ex}
\subsection{Correctness criteria}
\label{sec:labels}
\vspace{-0.6ex}

We evaluate candidates with two automatic criteria. \emph{Reference matching}
reimplements the matcher of \citet{ferrando2025entity}, using fuzzy name
matching and relation-specific numerical tolerances; a reference match need not
constitute a correct factual answer.

Our primary correctness criterion, the \emph{stricter criterion}, was developed on
an earlier cohort and fixed before prospective generation. It requires contiguous
name matches, rejects local negations and responses formatted as option lists, even when they end by choosing an answer, and applies
stricter numerical checks. Both criteria label every candidate, allowing the
criterion to vary while questions, candidates, scores, and selections remain
fixed. Full definitions appear in Appendix~\ref{app:scoring}.
We audit both criteria with blinded human annotation of $400$ candidates.
Section~\ref{sec:selection-label-sensitivity} further adds semantic
labels for all candidates from GPT-5.6 Sol \citep{openai2026gpt56} and human
labels for every output on which the two selectors disagree; protocols are
given in Appendices~\ref{app:human-audit} and~\ref{app:semantic}.

\vspace{-0.6ex}
\subsection{Recall-readout evaluation}
\label{sec:recall-eval}
\vspace{-0.6ex}
For each model, entity recall is the fraction of pre-specified relations answered
correctly under greedy decoding, scored by reference matching. We test whether a readout $S$ adds predictive
information beyond measured exposure and identity features by comparing
\[
M_0=
\text{exposure}+\text{name likelihood}+\text{name length}
+\text{type}+\text{relation},
\qquad
M_1=M_0+S.
\]
Exposure is the entity's Wikidata sitelink count and its 2024 Wikipedia
pageviews, each entered as $\log(1+x)$ with a missingness indicator; the
identity features are the name's mean token log-probability, its length in
characters and tokens, the entity type, and the relation. Both are
L2-regularized logistic models ($C=1$) with identical preprocessing, predicting
whether each relation query matches the reference. We average relation-level losses within each type and
then equally across types, reporting the type-macro log-loss improvement
\[
\Delta L=L(M_0)-L(M_1),
\]
where $\Delta L>0$ indicates improved held-out recall prediction.

Recall intervals use $10{,}000$ joint-refit entity-bootstrap draws, resampling
fit and confirmation entities within type. Selection comparisons and the
conversion interval use $20{,}000$ paired entity-clustered bootstrap draws, and
the primary candidate-budget curves
average $400$ without-replacement subsamples per saved pool. Appendix~\ref{app:provenance} lists which analyses were fixed before prospective
generation.

\vspace{-0.6ex}
\section{Recall Prediction Beyond Exposure}
\label{sec:construct}
\vspace{-0.6ex}

\begin{figure}[t]
\centering
\captionsetup[subfigure]{font=small,labelfont=bf,justification=raggedright,
singlelinecheck=false,skip=4pt}
\begin{subfigure}[t]{0.485\linewidth}
\centering
\includegraphics[width=0.8\linewidth]{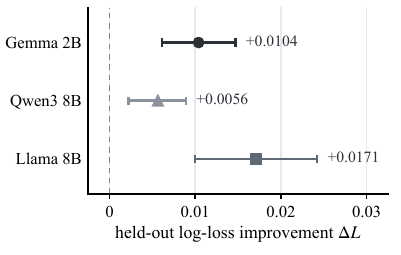}
\caption{\textbf{Across model families.}}
\end{subfigure}\hfill
\begin{subfigure}[t]{0.485\linewidth}
\centering
\includegraphics[width=0.8\linewidth]{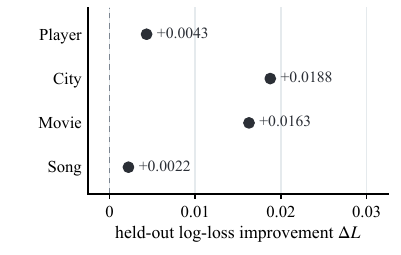}
\caption{\textbf{Across Gemma entity types.}}
\end{subfigure}
\caption{\textbf{Recall information beyond exposure.} $\Delta L$ is the
held-out log-loss reduction from adding the readout. Panel (a) shows type-macro
gains with $95\%$ joint-refit entity-bootstrap intervals; (b) shows Gemma by type.}
\label{fig:recall-information}
\vspace{-0.8ex}
\end{figure}

We first test whether internal states carry information about factual recall
beyond measured exposure and identity features. Across Gemma, Qwen3, and Llama,
adding a pre-generation readout to the same baseline covariates improves
held-out recall prediction in both log loss and Brier score
(Figure~\ref{fig:recall-information}; Appendix~\ref{app:recall-tables},
Table~\ref{tab:construct-confirmation}). Section~\ref{sec:know-choose} then
applies the three readouts to the selection questions.

\vspace{-0.6ex}
\subsection{Cross-model recall prediction}
\vspace{-0.6ex}
Gemma provides the prospective confirmation. Across $3{,}840$ entities and $16{,}320$ relation queries, adding the layer-$16$ class-mean readout improves type-macro log loss by $0.0104$ ($3.5\%$ of baseline) and Brier score by $0.0038$ ($4.2\%$), with both $95\%$ CIs above zero and positive effects across all four entity types (Appendix~\ref{app:recall-tables}, Table~\ref{tab:construct-types}). The result extends to Qwen3-8B and Llama~3.1 8B-Instruct: the readout improves log loss by $0.0056$ ($1.7\%$) and $0.0171$ ($3.9\%$), and Brier score by $0.0020$ ($1.9\%$) and $0.0064$ ($4.6\%$), respectively, again with both CIs above zero. Qwen3 improves three of four entity-type point estimates and Llama all four. Qwen3 uses an independent entity cohort, while Llama uses Gemma's entity set with a separately fitted readout. In Gemma and Llama, type-preserving label permutations yield near-zero improvements, while the reference answer's own likelihood improves log loss three to five times more than the readout (Appendices~\ref{app:reproduction} and~\ref{app:llama}). Qwen3 meets its permutation control only under a limit raised after its first cohort exceeded the pre-specified one; on the second cohort, the median permutation improvement, $0.0015$, also exceeds the original limit of $0.001$, and the $95$th percentile reaches $0.0025$, against an effect of $0.0056$ (Appendix~\ref{app:qwen}).

\vspace{-0.6ex}
\subsection{Readouts predict coverage more than selection success}
\label{sec:know-choose}
\vspace{-0.6ex}

\begin{figure}[t]
\centering
\captionsetup[subfigure]{font=small,labelfont=bf,justification=raggedright,
singlelinecheck=false,skip=4pt}
\begin{subfigure}[t]{0.32\linewidth}
\centering
\includegraphics[width=\linewidth]{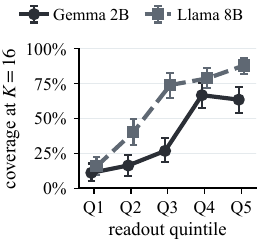}
\caption{\textbf{Across questions, the readout predicts coverage.}}
\end{subfigure}\hfill
\begin{subfigure}[t]{0.32\linewidth}
\centering
\includegraphics[width=\linewidth]{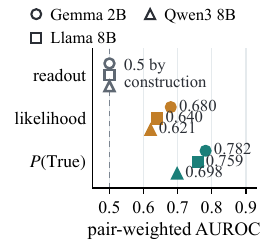}
\caption{\textbf{Within a question, verification ranks candidates best.}}
\end{subfigure}\hfill
\begin{subfigure}[t]{0.32\linewidth}
\centering
\includegraphics[width=\linewidth]{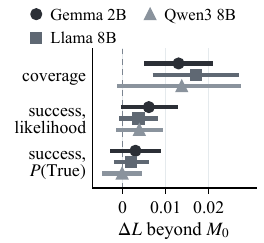}
\caption{\textbf{Beyond the baselines, it informs coverage more than
selection in Gemma and Llama.}}
\end{subfigure}
\caption{\textbf{Pre-generation readout across and within questions.}
(a)~$K=16$ coverage by readout quintile under the stricter criterion ($521$
Gemma questions after excluding entities used to fit the direction; $598$ Llama).
(b)~Pair-weighted AUROC on mixed questions ($199$ Gemma, $302$ Llama, $168$
Qwen3). (c)~Held-out log-loss improvement from adding the readout to
$M_0$, for coverage and for plurality success on covered questions ($192$,
$355$, and $244$). Intervals are $95\%$ entity-clustered.}
\label{fig:readout-coverage}
\vspace{-0.8ex}
\end{figure}

We apply the three readouts, without refitting, to predict whether each $K=16$ candidate pool contains a correct answer (Section~\ref{sec:selection}). After excluding $20$ Gemma questions whose entities were used to fit its direction, $521$ Gemma, $598$ Llama, and $600$ Qwen3 questions remain. Coverage AUROC is $0.7815$, $0.8165$, and $0.6898$, respectively; from the lowest to highest readout quintile, coverage rises from $11.1\%$ to $63.5\%$ in Gemma, $15.7\%$ to $87.9\%$ in Llama, and $27.5\%$ to $72.5\%$ in Qwen3 (Figure~\ref{fig:readout-coverage}a; Appendix~\ref{app:readout-coverage}). Controlling for entity familiarity and relation difficulty with entity-disjoint cross-fitting and a leave-one-entity-out common-answer prior, adding the readout improves held-out log loss by $0.013$ in Gemma ($95\%$ CI $[0.005,0.021]$) and $0.017$ in Llama ($[0.007,0.027]$). Qwen3's improvement over all questions, $0.0137$ ($[-0.0012,0.0276]$), includes zero but rises to $0.0246$ ($[0.0050,0.0468]$) in the six low-prior relations (Table~\ref{tab:readout-nested}). Because the readout is computed before generation, it scores the question rather than individual candidates and has within-question AUROC $0.5$ by construction (Figure~\ref{fig:readout-coverage}b). In Gemma and Llama, it improves AUROC by $0.040$ and $0.041$ for coverage but by only $0.006$--$0.015$ for plurality success, and its log-loss improvements for success are small, with intervals that include zero; Qwen3 shows no clear separation. Semantic labels yield the same pattern (Appendix~\ref{app:semantic}). The readout predicts whether sampling will expose a correct answer but says little about whether that answer will be selected. The SAE latents of \citet{ferrando2025entity}, whose unknown-side refusal steering we reproduce, predict recall less well than the class-mean readout and add little beyond exposure, type, and relation in detecting wrong chat answers (Appendix~\ref{app:sae}); relation identity alone reaches within-entity AUROC $0.93$, so we compare candidates within a question rather than within an entity (Appendix~\ref{sec:resolution}).

\vspace{-0.6ex}
\section{Converting Candidates into Final Answers}
\label{sec:selection}
\vspace{-0.6ex}

The main selection results use the prospectively specified $600$-question Gemma
cohort, with $541$ eligible questions and $451$ entity clusters; Llama, Qwen3,
Gemma~2 9B-IT and 27B-IT, and four further Qwen models are evaluated on the same
questions
(Section~\ref{sec:judge-choice}). We vary the candidate budget, compare the two
scores on fixed pools, locate what verification adds beyond generative baselines,
and test how the correctness criterion changes the measured gain.

\vspace{-0.6ex}
\subsection{Larger candidate budgets convert little of the added coverage}
\vspace{-0.6ex}
Selection here is plurality with a likelihood tie-break at $T=1$, scored by the
stricter criterion. The budget $K$ counts retained candidates, capped by
each question's pool of $8$--$16$, and all budgets are subsampled from the same
pools with questions and labels fixed.
From $K=1$ to $16$, coverage rises from $17.4\%$ to $37.7\%$,
while final accuracy increases only from $17.4\%$ to $23.1\%$
(Appendix~\ref{app:selection-tables}, Figure~\ref{fig:budget-curves} and
Table~\ref{tab:selection-budget}). At $K=16$, $204$ questions contain a correct
candidate but only $125$ yield a correct final answer, a conversion of
$61.3\%$ ($95\%$ CI $[54.6\%,67.9\%]$) that leaves $79$ covered questions
unconverted. Only $28.0\%$ of the coverage added from $K=1$ to $16$ becomes
additional final accuracy. Raising the sampling temperature has the same effect: in a separate run with fresh
samples for the same questions, moving from $T=0.3$ to $T=1$ raises coverage at
$K=16$ from $27.98\%$ to $36.45\%$ but likelihood accuracy only from $21.31\%$ to
$21.97\%$ (Appendix~\ref{app:measured-runtime}; Figure~\ref{fig:temperature}).

\vspace{-0.6ex}
\subsection{Explicit verification improves ranking and selection}
\label{sec:verification}
\vspace{-0.75ex}
We compare likelihood and verification on the same $6{,}864$ candidates from $541$ questions under the stricter criterion. Verification improves both within-question ranking and final selection (Appendix~\ref{app:selection-tables}, Table~\ref{tab:ptrue}). On the $199$ mixed questions, pair-weighted AUROC rises from $0.680$ to $0.782$ ($+0.102$, $95\%$ CI $[0.056,0.149]$). Under argmax selection, verification returns $145$ correct answers versus $116$ for mean log-likelihood, a $5.36$-point gain ($[2.56,8.19]$); as a plurality tie-breaker, it returns $152$ versus $125$, a $4.99$-point gain ($[2.7,7.3]$). Among the $204$ covered questions, plurality conversion rises from $61.3\%$ to $74.5\%$, closing about a third of the gap to full conversion (Appendix~\ref{app:selection-tables}). The gain persists across four prompt wordings ($3.33$--$4.99$ points), confidence-weighted voting ($4.44$--$5.91$), and an independent $K=16$ run without top-$k$ truncation ($4.34$ points, $[2.85,5.89]$; Appendices~\ref{app:prompt-robustness}, \ref{app:weighted-voting}, and~\ref{app:measured-runtime}). In the same run, lowering the temperature to $0.3$ shrinks the gain from $4.34$ to $1.68$ points (difference $2.66$, $[1.22,4.12]$; Figure~\ref{fig:temperature}), so what verification adds depends on the candidates that decoding produces. Greedy decoding reaches $27.54\%$ versus $28.10\%$ for verification plurality. The two scores also make complementary errors: under argmax, verification finds $45$ correct answers missed by likelihood, while likelihood finds $16$ missed by verification, giving an oracle accuracy of $29.8\%$ (Figure~\ref{fig:development-selection}b).

\vspace{-0.8ex}
\subsection{Verification ranks better in seven of nine models and after text filtering in Gemma and Llama}
\label{sec:judge-choice}
\vspace{-0.75ex}

Table~\ref{tab:models} compares likelihood and verification across models. Verification ranks candidates better in all models except Gemma~2 27B-IT and Qwen3.5-2B, and its plurality gain is largest in Gemma, where likelihood leaves room for improvement and verification also ranks candidates better. Likelihood converts $61.27\%$ of covered Gemma questions, versus $73.24\%$ in Llama and $81.88\%$/$81.63\%$ in Gemma~2 9B-IT/27B-IT, and ranks a correct answer first on $55.8\%$ of mixed questions versus $68.5\%$ in Llama (Appendices~\ref{app:rank-analysis} and~\ref{app:gemma27b}), while verification raises pair-weighted AUROC by $0.102$. Room for improvement alone is not enough: in Qwen3.5-2B, likelihood converts only $58.65\%$ of covered questions, but verification barely changes ranking ($+0.008$) and lowers plurality conversion to $50.48\%$. Option lists explain part of the Gemma gain: likelihood selects one on $61$ Gemma questions versus $11$ for verification, yet after removing option lists, negations, and refusals, verification retains $3.52$ of its $4.99$-point gain ($95\%$ CI $[1.57,5.47]$) and still ranks better in both Gemma and Llama (Appendix~\ref{app:rank-analysis}). Llama's $1.67$-point gain is less robust across prompt variants and aggregation rules (Appendix~\ref{app:llama-selection}). Across five Qwen models from 2B to 27B, coverage ranges from $19.70\%$ to $48.58\%$, while likelihood converts $58.20\%$--$64.34\%$ of covered questions (Figure~\ref{fig:qwen}a; Appendix~\ref{app:qwen-family}). Verification changes argmax accuracy by $-1.80$ points in Qwen3.5-2B and $+2.30$ in Qwen3.5-4B, and raises plurality accuracy by $3.34$ and $3.40$ points in the two 27B models, where likelihood leaves similar room ($61.86\%$ and $60.00\%$) and verification ranks candidates better ($+0.180$ and $+0.149$; Appendix~\ref{app:qwen-family}).

\vspace{-0.8ex}
\subsection{The gain beyond generative baselines depends on the entity}
\label{sec:rescoring}
\begin{figure}[htb]
\centering
\captionsetup[subfigure]{font=small,labelfont=bf,justification=raggedright,
singlelinecheck=false,skip=4pt}
\begin{subfigure}[t]{0.33\linewidth}
\centering
\includegraphics[width=\linewidth]{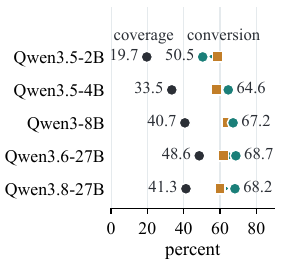}
\caption{\textbf{Verification raises conversion except in Qwen3.5-2B.}}
\end{subfigure}\hfill
\begin{subfigure}[t]{0.32\linewidth}
\centering
\includegraphics[width=\linewidth]{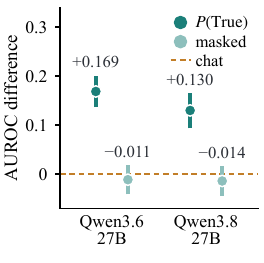}
\caption{\textbf{Ranking: masking the entity removes verification's advantage.}}
\end{subfigure}\hfill
\begin{subfigure}[t]{0.32\linewidth}
\centering
\includegraphics[width=\linewidth]{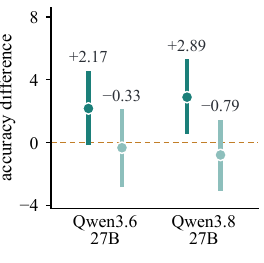}
\caption{\textbf{Selection: the advantage is two to three points.}}
\end{subfigure}
\caption{\textbf{Verification across recent Qwen models.} Stricter criterion, $16$ raw samples per question.
(a) Coverage and plurality conversion across five models (likelihood amber,
$P(\mathrm{True})$ teal).
(b,c) Ranking and selection differences relative to chat-template likelihood
for $P(\mathrm{True})$ and entity-masked $P(\mathrm{True})$ on Qwen3.6-27B
and Qwen3.8-27B. Error bars show paired $95\%$ intervals.}
\label{fig:qwen}
\end{figure}

\vspace{-0.75ex}
The main comparison scores likelihood on the prompt used for sampling but verification under the chat template, so part of the gap may reflect the prompt format. We rescore the same candidates with summed likelihood, chat-template likelihood, and relation-prior PMI, which subtracts the likelihood of the same candidate when the entity is replaced by \emph{this \{type\}} (Appendix~\ref{app:rescoring}). Under semantic labels, both summed and chat-template likelihood close part of the gap, whereas PMI performs worse than mean log-likelihood (Figure~\ref{fig:decomposition}a): chat-template likelihood recovers $10$ of verification's $21$ additional correct selections, but verification remains $2.03$ points higher ($95\%$ CI $[0.16,3.92]$). Verification's gain over mean log-likelihood is concentrated in the four high-prior relations, $7.86$ points ($[3.55,12.10]$) versus $0.96$ in the other six; chat-template likelihood closes most of it there, leaving verification $3.06$ points higher in the high-prior relations ($[0.12,5.99]$) against $1.28$ in the others (Figure~\ref{fig:decomposition}b). Entity masking removes most of the remainder: on $229$ high-prior questions, masked $P(\mathrm{True})$ is only $0.44$ points above chat-template likelihood ($[-1.55,2.49]$) and $2.62$ below unmasked $P(\mathrm{True})$ ($[0.11,5.18]$), reducing verification's net advantage from seven correct selections to one (Figure~\ref{fig:decomposition}b,c). The same pattern appears in two 27B Qwen models: verification improves pair-weighted AUROC over chat-template likelihood by $0.13$--$0.17$ but plurality accuracy by only $2.2$--$2.9$ points, and masking the entity reduces both ranking and accuracy to about the level of chat-template likelihood (Figure~\ref{fig:qwen}b,c). In Gemma's high-prior relations and in both 27B Qwen models, verification's advantage over chat-template likelihood depends on the entity.

\begin{figure}[htb]
\centering
\captionsetup[subfigure]{font=small,labelfont=bf,justification=raggedright,
singlelinecheck=false,skip=4pt}
\begin{subfigure}[t]{0.29\linewidth}
\centering
\includegraphics[width=\linewidth]{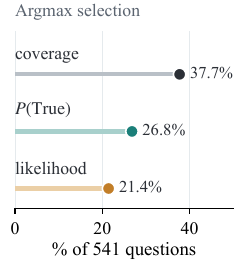}
\caption{\textbf{Verification converts more of the same coverage.}}
\end{subfigure}\hfill
\begin{subfigure}[t]{0.25\linewidth}
\centering
\includegraphics[width=\linewidth]{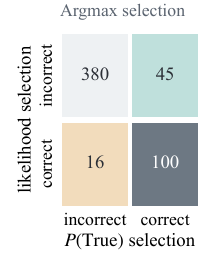}
\caption{\textbf{Each score recovers answers the other misses.}}
\end{subfigure}\hfill
\begin{subfigure}[t]{0.42\linewidth}
\centering
\includegraphics[width=\linewidth]{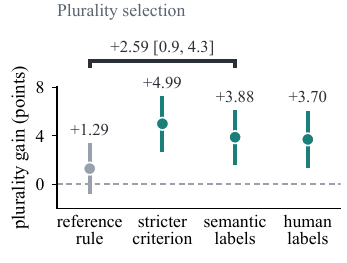}
\caption{\textbf{Reference matching hides most of the gain.}}
\end{subfigure}
\caption{\textbf{From coverage to final answers.} Gemma, $541$
questions, $K=16$. (a,\,b)~Argmax selection under the stricter criterion; in (b),
$161$ of the $204$ covered questions have at least one correct selection.
(c)~Plurality gain of $P(\mathrm{True})$ over mean log-likelihood under four
labelings of the same selected outputs, with paired $95\%$ intervals; the bracket
gives the difference between semantic labels and reference matching.}
\label{fig:development-selection}
\vspace{-0.6ex}
\end{figure}

\vspace{-0.6ex}
\subsection{Reference matching hides most of the verification gain}
\label{sec:selection-label-sensitivity}
\vspace{-0.6ex}
Under reference matching, verification's plurality gain over mean log-likelihood is only $1.29$ points, versus $4.99$ under the stricter criterion, despite nearly unchanged coverage (Figure~\ref{fig:development-selection}c; Appendix~\ref{app:selection-tables}). We label all $6{,}864$ Gemma candidates semantically with GPT-5.6 Sol \citep{openai2026gpt56} using a rubric fixed before any selector was compared under these labels, and obtain human labels from three authors for every output on which the selectors disagree. GPT-5.6 Sol agrees with human labels on $93.2\%$ of sampled candidates (Cohen's $\kappa=0.84$). The 400-candidate audit favors the stricter criterion on the disputed selections, although on $100$ uniformly sampled questions its two annotators find argmax gains of $3$ and $-1$ questions, against $8$ under the stricter criterion (Appendices~\ref{app:human-audit} and~\ref{app:semantic}). Both labelings recover most of the stricter criterion's gain: verification improves plurality accuracy by $3.88$ points ($[1.6,6.1]$) under semantic labels and $3.70$ ($[1.31,6.08]$) under human labels, so reference matching hides $2.59$ points ($[0.9,4.3]$; Table~\ref{tab:semantic-endpoints}). The discrepancy arises mainly because reference matching credits option lists containing the reference answer: relative to semantic labels, it mislabels $3.51\%$ of likelihood's argmax selections versus $0.55\%$ of verification's, and $18$ of the $19$ errors on likelihood's selections come from such lists (Table~\ref{tab:reference-errors}).

\vspace{-0.6ex}
\section{Limitations}
\label{sec:limitations}
\vspace{-0.6ex}

The selection comparisons in Section~\ref{sec:selection} use Gemma, Llama, Qwen3,
Gemma~2 9B and 27B, and four further Qwen models on short English factual questions with filtered candidate pools. In Qwen3,
the readout's improvement over the baselines of
Section~\ref{sec:know-choose} has an interval that includes zero over all questions. The primary correctness criterion is an automatic rule. The semantic labels were
added after the prospective analysis, and the human labels come from three of the
authors. Recall
labels in Section~\ref{sec:construct} use reference matching, and the coverage
analyses in Section~\ref{sec:know-choose} and the rescoring analyses in
Section~\ref{sec:rescoring} are secondary analyses of saved candidate pools.

\vspace{-0.6ex}
\section{Conclusion}
\vspace{-0.6ex}

Sampling produces correct answers that the model does not return. Pre-generation
readouts predict factual recall and which questions sampling will cover in Gemma,
Qwen3, and Llama, but say little about whether a covered question ends with a
correct answer. Asking the model whether a candidate is correct ranks a question's
candidates better than its likelihood in all three models. The plurality gain over
mean log-likelihood is largest in Gemma, where likelihood converts only $61.3\%$ of
covered questions and verification also ranks candidates better; Qwen3.5-2B leaves
as much room but, without a ranking advantage, gains nothing. Against likelihood
scored under the chat template, $2.03$ points remain in Gemma. Masking the entity removes most of verification's remaining
advantage in Gemma's high-prior relations and in two 27B Qwen models, where a large
ranking advantage moves final accuracy by only two to three points. Reference
matching credits option lists that contain the reference answer and so hides most
of the gain that semantic and human labels confirm.
Together, these results expose two distinct bottlenecks in factual generation: whether a correct answer becomes available, and whether it survives competition among the alternatives that are generated. For Gemma and Llama, entity-level readouts carry recall information beyond measured exposure and identity features, but cannot resolve relation- or answer-specific correctness. Finally, recall-oriented reference matching systematically credits option lists favored by likelihood, hiding selection differences that semantic and human labels confirm.

\newpage
\subsection*{AI use statement}

We used GPT-5.6 Sol \citep{openai2026gpt56} for annotation, and Claude Opus 5.5
for manuscript polishing and experiment code development. All
AI-assisted content and analyses were reviewed by the authors.

\subsection*{Ethics statement}

The study uses public Wikidata entities \citep{vrandecic2014wikidata}, open-weight models \citep{gemmateam2024gemma2,yang2025qwen3,grattafiori2024llama3,qwen35,qwen36,qwen38}, and open sparse autoencoders \citep{lieberum2024gemmascope,deng2026qwenscope}. Model outputs
may contain factual errors about real public figures. We report aggregate
results and selected illustrative examples. The authors performed all human annotation: two authors labeled the initial
$400$-candidate audit (Appendix~\ref{app:human-audit}), and three authors produced
the human labels in Appendix~\ref{app:semantic}.

\subsection*{Reproducibility statement}

Section~\ref{sec:setup} and Appendices~\ref{app:reproduction}--\ref{app:scoring},
\ref{app:llama-selection}--\ref{app:measured-runtime}, and
\ref{app:qwen-selection}--\ref{app:prompts} specify model
revisions, prompts, decoding settings, correctness rules, eligibility filters,
and bootstrap procedures. We will release the code, frozen protocols, candidate
pools, scores, and labels.

\bibliographystyle{plainnat}
\bibliography{main}

\clearpage
\appendix
\section{Additional Selection Results}
\label{app:selection-tables}

Table~\ref{tab:ptrue} gives ranking and selection on the fixed Gemma candidates of
Section~\ref{sec:verification}; the tables after it cover candidate budgets,
complementarity, and correctness criteria on the same cohort.

\begin{table}[H]
\centering
\small
\setlength{\tabcolsep}{5pt}
\caption{\textbf{Ranking and selection on fixed candidates.} Stricter criterion,
$541$ questions, $T=1$, $K=16$. AUROC is pair-weighted over $199$ mixed
questions; conversion divides accuracy by coverage ($204/541$), using plurality accuracy
for the scored rows. Unscored plurality breaks ties by draw order, averaged over
$400$ random orderings of each pool. The oracle and greedy rows return one answer
per question, so their accuracy is listed under Argmax; the oracle returns a
correct candidate whenever one exists, so its conversion is $100\%$ by definition.
Greedy is a single-answer baseline on the same questions.}
\label{tab:ptrue}
\begin{tabular}{lrrrr}
\toprule
Method & Pair-weighted AUROC & Argmax selection & Plurality selection & Conversion \\
\midrule
Mean log-likelihood & $0.6798$ & $21.44\%$ & $23.11\%$ & $61.27\%$ \\
Summed log-likelihood & $0.6870$ & $24.40\%$ & $24.58\%$ & $65.20\%$ \\
$P(\mathrm{True})$ & $\mathbf{0.7820}$ & $\mathbf{26.80\%}$ & $\mathbf{28.10\%}$ & $\mathbf{74.51\%}$ \\
Plurality & --- & --- & $21.15\%$ & $56.08\%$ \\
\midrule
Oracle & --- & $37.71\%$ & --- & $100\%$ \\
\midrule
\multicolumn{5}{l}{\emph{Single-answer decoding}} \\
Greedy & --- & $27.54\%$ & --- & --- \\
\bottomrule
\end{tabular}
\end{table}

\begin{table}[H]
\centering
\small
\setlength{\tabcolsep}{5pt}
\caption{\textbf{Coverage grows faster than accuracy.}
$541$ eligible questions from the fresh $600$-question cohort, $T=1$,
stricter criterion, plurality with likelihood tie-breaking. $K$ is a retained-pool
budget, capped at the pool size. $R_K=A_K/C_K$ measures conversion.}
\label{tab:selection-budget}
\begin{tabular}{rrrr}
\toprule
$K$ & Coverage $C_K$ & Accuracy $A_K$ & Conversion $R_K$ \\
\midrule
$1$  & $0.1743$ & $0.1743$ & $1.0000$ \\
$2$  & $0.2389$ & $0.1952$ & $0.8171$ \\
$4$  & $0.2960$ & $0.2088$ & $0.7055$ \\
$8$  & $0.3482$ & $0.2216$ & $0.6364$ \\
$16$ & $0.3771$ & $0.2311$ & $0.6127$ \\
\bottomrule
\end{tabular}
\end{table}

\begin{figure}[H]
\centering
\begin{minipage}[c]{0.47\linewidth}
\centering
\includegraphics[width=\linewidth,pagebox=cropbox]{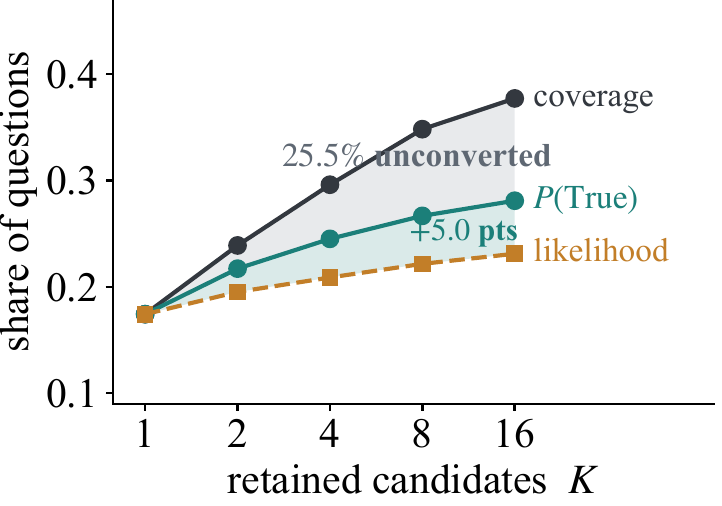}
\end{minipage}\hfill
\begin{minipage}[c]{0.49\linewidth}
\caption{\textbf{Little of the added coverage is converted.} The $541$ prospective Gemma questions
under the stricter criterion. Black shows coverage; teal and dashed amber show
plurality accuracy with $P(\mathrm{True})$ and likelihood. At $K=16$, verification
raises conversion from $61.3\%$ to $74.5\%$, improving accuracy by $4.99$ points
while $25.5\%$ of covered questions remain unconverted.}
\label{fig:budget-curves}
\end{minipage}
\end{figure}

\paragraph{Complementary choices.} Under argmax selection in Gemma, verification
contributes $45$ successes that likelihood misses, while likelihood contributes $16$
in the other direction (Table~\ref{tab:complementarity}). Their union contains
correct selections for $161$ of the $204$ covered questions
(Figure~\ref{fig:development-selection}b), and an oracle choosing between their two
outputs would reach $29.8\%$. Rank-sum fusion, which returns the candidate with the
highest sum of its within-question ranks under the two scores, reaches $25.5\%$,
against $26.8\%$ for verification.

\begin{table}[H]
\centering
\small
\setlength{\tabcolsep}{8pt}
\caption{\textbf{Likelihood and verification recover different correct answers.}
Argmax outcomes partition the same $541$ questions under the stricter criterion.
The oracle uses labels to choose between the two selectors' outputs, so it
measures their combined opportunity.}
\label{tab:complementarity}
\begin{tabular}{lr}
\toprule
Selection outcome & Questions \\
\midrule
Both select a correct answer & $100$ \\
Only likelihood selects a correct answer & $16$ \\
Only $P(\mathrm{True})$ selects a correct answer & $45$ \\
Neither selects a correct answer & $380$ \\
\midrule
Oracle over the two selections & $161/541=29.8\%$ \\
Candidate coverage & $204/541=37.7\%$ \\
Oracle conversion & $161/204=78.9\%$ \\
\bottomrule
\end{tabular}
\end{table}

\begin{table}[H]
\centering
\small
\setlength{\tabcolsep}{4pt}
\caption{\textbf{Selection gains under the two automatic criteria.}
Same $541$ questions, candidates, scores, and selected outputs; only correctness
labels change. Accuracies are percentages; differences and paired $95\%$
entity-clustered intervals ($20{,}000$ draws) are percentage points.}
\label{tab:label-sensitivity}
\begin{tabular}{llrrrl}
\toprule
Criterion & Selection & Likelihood & $P(\mathrm{True})$ & Difference & 95\% CI \\
\midrule
Reference & Argmax & $26.62$ & $27.36$ & $+0.74$ & $[-1.71,3.19]$ \\
Stricter & Argmax & $21.44$ & $26.80$ & $\mathbf{+5.36}$ & $[2.56,8.19]$ \\
\midrule
Reference & Plurality & $26.99$ & $28.28$ & $+1.29$ & $[-0.8,3.4]$ \\
Stricter & Plurality & $23.11$ & $28.10$ & $\mathbf{+4.99}$ & $[2.7,7.3]$ \\
\bottomrule
\end{tabular}
\end{table}

\begin{figure}[H]
\centering
\includegraphics[width=0.55\linewidth]{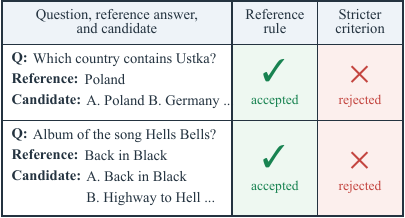}
\caption{\textbf{Reference matching credits option lists.} Two option lists
that contain the reference answer: reference matching accepts them and the
stricter criterion rejects them. Figure~\ref{fig:development-selection}c shows the resulting plurality gains under each criterion.}
\label{fig:criterion-measurement}
\end{figure}

Under reference matching, verification's argmax and plurality gains over mean
log-likelihood are $0.74$ and $1.29$ points, with both intervals including zero;
under the stricter criterion, they are $5.36$ and $4.99$ points, with both
intervals above zero (Table~\ref{tab:label-sensitivity}). Coverage changes little
($210/541$ to $204/541$), while the paired criterion-by-evaluator interaction is
$4.62$ points for argmax ($[2.91,6.53]$) and $3.70$ for plurality ($[2.2,5.3]$).
Option-list credit fully explains this attenuation: among outputs accepted only by
reference matching, all $28$ likelihood selections and all $3$ verification
selections are option lists; their differential credit accounts for the argmax
attenuation, and under plurality $21$ versus $1$ such selections account for the
full $3.70$-point change. \citet{kamalloo2023evaluating} document the converse
failure, in which lexical matching rejects correct free-form answers; here
reference matching credits option lists that contain the reference answer, as in
the two examples of Figure~\ref{fig:criterion-measurement}. Ranking shows the same
attenuation: the AUROC gap between verification and likelihood is $0.102$ under
the stricter criterion and $0.022$ under reference matching, where pair-weighted
AUROC rises from $0.711$ to $0.733$ (Table~\ref{tab:semantic-endpoints}).

\paragraph{Notes on Table~\ref{tab:ptrue}.}
Table~\ref{tab:ptrue} uses the saved scores and labels. The unscored Plurality
row breaks ties by the first draw over $400$ random orderings of each pool; it
reaches $21.15\%$ accuracy and converts $56.08\%$ of covered questions.
Its conversion column uses plurality; argmax conversion rises from $56.9\%$ to $71.1\%$.
Among the $204$ covered questions, unconverted questions fall from $79$ to $52$
under plurality, closing $34.2\%$ of that gap, and argmax closes $33.0\%$; under
plurality, verification gains $35$ correct selections and loses $8$. The candidate oracle in Table~\ref{tab:ptrue}
can select from the entire pool, whereas the oracle over the two selections in
Table~\ref{tab:complementarity} can only choose between them.

\paragraph{Filtered likelihood.} Table~\ref{tab:extra-baselines} applies the
text filter of Appendix~\ref{app:rank-analysis} before likelihood selection
on the same $541$ questions; every question keeps at least one candidate after
filtering.

\begin{table}[H]
\centering
\small
\setlength{\tabcolsep}{5pt}
\caption{\textbf{Likelihood after the text filter.} Stricter criterion,
$541$ questions, $T=1$, $K=16$; conversion is plurality accuracy divided by
coverage.}
\label{tab:extra-baselines}
\begin{tabular}{lrrr}
\toprule
Method & Argmax selection & Plurality selection & Conversion \\
\midrule
Mean log-likelihood, filtered & $25.32\%$ & $25.69\%$ & $68.14\%$ \\
\bottomrule
\end{tabular}
\end{table}

\paragraph{Development sampling cohort.} The same comparison was first run on this cohort (Table~\ref{tab:cohorts}), on which the stricter criterion was developed: $152$ name-valued questions over $128$ entities at $T=1$ and $K=16$. Under
the stricter criterion, $53$ questions ($34.9\%$) are covered, and $28.2\%$ of the
coverage added from $K=1$ to $16$ becomes plurality accuracy, against $28.0\%$ in the
prospective cohort. Over $51$ mixed questions and $1{,}257$ pairs, verification
raises query-macro AUROC from $0.6619$ to $0.8003$ ($+0.1384$, $[0.0513,0.2250]$)
and pair-weighted AUROC from $0.6611$ to $0.7749$ ($+0.1138$, $[0.0190,0.2097]$).
As a plurality tie-breaker it returns $40$ correct answers against $32$ for
likelihood ($+5.26$ points, $[0.67,10.07]$), raising conversion from $60.4\%$ to
$75.5\%$; under argmax it returns $37$ against $29$ ($+5.26$, $[-0.66,11.39]$). The
two scores pick the same candidate on $9.2\%$ of questions: $22$ questions are
answered correctly by both, $7$ by likelihood only, $15$ by verification only, and
$108$ by neither, so an oracle over the two reaches $28.9\%$. Under reference
matching, the two selectors return the same number of correct answers, $38$ under
argmax and $41$ under plurality. These intervals use $10{,}000$ entity-bootstrap
draws.

\section{Recall Study Tables}
\label{app:recall-tables}

These tables give the loss and type-specific measurements supporting
Section~\ref{sec:construct}, and the two sparse autoencoder latents.

\begin{table}[H]
\centering
\small
\setlength{\tabcolsep}{4pt}
\caption{\textbf{Internal recall information extends across model families.}
Each model is evaluated on $3{,}840$ entities, including a $1{,}280$-entity
confirmation role. Gemma and Llama use matched cohorts and layer-$16$ and
layer-$15$ class-mean readouts; Qwen3 uses a separate cohort and fixed
layer-$23$ feature under its amended protocol (Appendix~\ref{app:qwen}).
Losses and improvements are equal-weighted type
macros; intervals use $10{,}000$ joint-refit entity-bootstrap draws.}
\label{tab:construct-confirmation}
\begin{tabular}{llrrrr}
\toprule
Model & Metric & $M_0$ & $M_1$ & Improvement & 95\% CI \\
\midrule
Gemma~2 2B-IT & Log loss & $0.295885$ & $0.285492$ & $+0.010393$ & $[0.006103,0.014701]$ \\
 & Brier & $0.088757$ & $0.084993$ & $+0.003764$ & $[0.002253,0.005272]$ \\
\midrule
Qwen3-8B & Log loss & $0.328104$ & $0.322458$ & $+0.005646$ & $[0.002218,0.008922]$ \\
 & Brier & $0.100192$ & $0.098240$ & $+0.001952$ & $[0.000765,0.003144]$ \\
\midrule
Llama~3.1 8B-Instruct & Log loss & $0.436713$ & $0.419618$ & $+0.017095$ & $[0.009969,0.024223]$ \\
 & Brier & $0.139928$ & $0.133543$ & $+0.006385$ & $[0.003911,0.008907]$ \\
\bottomrule
\end{tabular}
\end{table}

\begin{table}[H]
\centering
\small
\caption{\textbf{Recall improvement by entity type.}
Type-specific point estimates for all three recall studies. The readout
improves both scores in all four types for Gemma and Llama and in three for Qwen3.}
\label{tab:construct-types}
\setlength{\tabcolsep}{4pt}
\begin{tabular}{lrrrrrr}
\toprule
 & \multicolumn{2}{c}{Gemma~2 2B-IT} & \multicolumn{2}{c}{Qwen3-8B} & \multicolumn{2}{c}{Llama~3.1 8B-Instruct} \\
\cmidrule(lr){2-3}\cmidrule(lr){4-5}\cmidrule(lr){6-7}
Type & $\Delta$ log loss & $\Delta$ Brier & $\Delta$ log loss & $\Delta$ Brier & $\Delta$ log loss & $\Delta$ Brier \\
\midrule
Player & $+0.004327$ & $+0.000986$ & $-0.001800$ & $-0.000648$ & $+0.021228$ & $+0.009470$ \\
City   & $+0.018755$ & $+0.008791$ & $+0.012650$ & $+0.005363$ & $+0.013850$ & $+0.003681$ \\
Movie  & $+0.016274$ & $+0.005028$ & $+0.009230$ & $+0.002342$ & $+0.019916$ & $+0.007289$ \\
Song   & $+0.002215$ & $+0.000252$ & $+0.002505$ & $+0.000751$ & $+0.013388$ & $+0.005102$ \\
\bottomrule
\end{tabular}
\end{table}

\begin{table}[H]
\centering
\small
\caption{\textbf{SAE latents also carry recall information.}
Improvements on the prospective Gemma recall cohort, with fitted models held fixed.}
\label{tab:construct-sparse}
\begin{tabular}{lcc}
\toprule
Readout & Log-loss improvement (95\% CI) & Brier improvement (95\% CI) \\
\midrule
Latent $7957$ & $+0.007234\ [0.003492,0.011051]$
              & $+0.002566\ [0.001304,0.003877]$ \\
$-\,$latent $11898$ & $+0.002124\ [0.000664,0.003566]$
                    & $+0.000492\ [0.000017,0.000952]$ \\
\bottomrule
\end{tabular}
\end{table}

\section{Development Ranking Tables}
\label{app:discrimination-tables}

These tables supplement Appendix~\ref{sec:resolution} and use the earlier same-question sample and the development sampling cohort (Appendix~\ref{app:scoring}), labeled by reference matching.

\begin{table}[H]
\centering
\small
\setlength{\tabcolsep}{4pt}
\caption{\textbf{Within-question ranking persists when the greedy answer is wrong.}
In the earlier same-question sample, likelihood ranks sampled answers that match the reference
against sampled answers that do not.
Intervals are entity-clustered.}
\label{tab:query-discrimination}
\begin{tabular}{p{0.45\linewidth}rrl}
\toprule
Comparison & Questions & Query-macro AUROC & 95\% CI \\
\midrule
Greedy answer matches: matching vs.\ non-matching samples & $155$ & $0.7569$ & $[0.7323,0.7829]$ \\
Greedy answer misses: matching vs.\ non-matching samples & $90$ & $0.6073$ & $[0.5685,0.6475]$ \\
\bottomrule
\end{tabular}
\end{table}

\begin{table}[H]
\centering
\small
\setlength{\tabcolsep}{5pt}
\caption{\textbf{Pooled and within-question ranking respond differently to decoding.}
The development sampling cohort is sampled at four temperatures (Appendix~\ref{app:scoring}). Statistics
use retained responses under reference matching; the mixed-question population
changes with temperature. Query-macro AUROC weights mixed questions equally.}
\label{tab:temperature}
\begin{tabular}{rrrrr}
\toprule
Temperature & Accuracy & Mixed questions & Query-macro AUROC & Pooled AUROC \\
\midrule
$0.3$ & $26.7\%$ & $38$  & $0.6312$ & $0.7729$ \\
$0.7$ & $24.4\%$ & $88$  & $0.6692$ & $0.7646$ \\
$1.0$ & $19.7\%$ & $109$ & $0.7055$ & $0.7429$ \\
$1.3$ & $15.2\%$ & $106$ & $0.7378$ & $0.7427$ \\
\bottomrule
\end{tabular}
\end{table}

\section{Study Populations}
\label{app:cohorts}

\begin{table}[H]
\centering
\small
\setlength{\tabcolsep}{4pt}
\caption{Study populations. Denominators are not pooled across studies.}
\label{tab:cohorts}
\begin{tabular}{p{0.23\linewidth}p{0.24\linewidth}p{0.43\linewidth}}
\toprule
Study & Model and population & Evaluation \\
\midrule
Prospective recall & Gemma~2 2B-IT; $3{,}840$ new entities &
Fit/calibration/confirmation roles of $1{,}280$ entities each \\
Recall replication & Qwen3-8B; $3{,}840$ new entities &
Separate fixed feature, cohort, and protocol \\
Recall replication & Llama~3.1 8B-Instruct; $3{,}840$ entities &
Gemma-matched cohort and roles; separate layer-$15$ class-mean direction \\
Earlier same-question sample & Gemma; $600$ questions &
$16$ samples at $T=1$, top-$p$ $0.95$; $245$ retained mixed questions \\
Development sampling cohort & Gemma; separate sample of $250$ questions & Stricter-criterion temperature cohorts: $120$ name-valued, $170$ answerable \\
$512$-entity chat cohort & Gemma~2 2B-IT and 2B; $512$ entities & $2{,}176$ human-labeled chat responses; paired pretrained and instruction-tuned recall study \\
Prospective selection & Gemma; $600$ fresh questions at $T=1$ &
$541$ eligible questions, $451$ entities, $6{,}864$ retained candidates \\
Matched-question selection & Llama~3.1 8B-Instruct; same $600$ questions &
$598$ eligible questions, $489$ entities, $9{,}349$ retained candidates \\
Matched-question selection & Qwen3-8B; same $600$ questions &
$600$ eligible questions, $489$ entities, $9{,}443$ retained candidates \\
Larger-model selection & Gemma~2 9B-IT; same $600$ questions &
$581$ eligible questions \\
Larger-model selection & Gemma~2 27B-IT; same $600$ questions &
$564$ eligible questions \\
Additional Qwen selection & Qwen3.5-2B, Qwen3.5-4B, Qwen3.6-27B, Qwen3.8-27B; same $600$ questions &
$528$, $565$, $599$, and $594$ eligible questions \\
Readouts on selection questions & Gemma, Llama, and Qwen3; selection questions &
$521$ Gemma questions ($435$ entities) after excluding direction-fit entities;
all $598$ eligible Llama questions; all $600$ Qwen3 questions, $244$ covered \\
Blinded human annotation audit & $400$ prospective candidates; two authors &
Both argmax selections on $100$ uniformly sampled questions, plus enriched examples \\
Human labels & Gemma prospective outputs; three authors &
$500$ outputs on which the selectors differ, $250$ random candidates, $200$ hard-strata candidates \\
\bottomrule
\end{tabular}
\end{table}

\section{Provenance}
\label{app:provenance}

Table~\ref{tab:provenance} lists the status of the analyses reported in the
paper. The full records will be released with the code.

\begin{table}[h]
\centering
\small
\setlength{\tabcolsep}{3pt}
\caption{Status of the reported analyses.}
\label{tab:provenance}
\begin{tabular}{p{0.44\linewidth}p{0.48\linewidth}}
\toprule
Analysis & Status \\
\midrule
Gemma recall confirmation & Protocol fixed before measurement; positive \\
Qwen3 independent-cohort recall & Positive under the amended permutation limit \\
Llama matched-cohort recall & Positive; pre-specified controls met \\
Gemma prospective selection & Protocol fixed before generation \\
Llama selection inputs & Same questions and fixed rules as Gemma \\
Qwen3 selection inputs & Same questions and fixed rules as Gemma \\
Gemma~2 9B and 27B selection inputs & Same questions and fixed rules as Gemma \\
Additional Qwen models & Post hoc; same questions and correctness criteria \\
Generative rescoring and entity masking & Secondary analysis of saved candidate pools \\
Sampling at $T=0.3$ & Secondary comparison; same design as the fresh $T=1$ samples \\
Confidence-weighted voting & Secondary comparison \\
Blinded human annotation audit & Two-author human annotations \\
Readout prediction of coverage and selection & Secondary analysis; readouts fixed, nested models cross-fitted \\
Semantic judge and human labels & Post hoc; protocol fixed before any selector was compared under these labels \\
SAE refusal steering & Reproduction of \citet{ferrando2025entity}; automatic refusal counts; known side post hoc \\
Latent $11{,}898$ on the $512$-entity chat cohort & Development analysis; human labels \\
\bottomrule
\end{tabular}
\end{table}

The Gemma recall confirmation uses a new $3{,}840$-entity role assignment, a
direct nested model, joint-refit bootstrap intervals, and controls computed
before outcomes.

\section{Gemma Recall Readout}
\label{app:reproduction}

\paragraph{Readout construction.} The class-mean direction is fitted on
$112$ entities, disjoint from the prospective recall cohort. At each candidate layer,
we standardize entity-final residual activations and form an equal-type
known-minus-unknown class-mean direction, normalized to unit length. Nested
leave-one-type-out selection chooses layer $16$. The direction, centering, and
scaling are fixed before prospective collection, and the same transformation is
applied to every new entity. The nested recall regressions use
L2-regularized logistic regression with $C=1$, \texttt{lbfgs}, at most
$5{,}000$ iterations, and tolerance $10^{-10}$.

\paragraph{Recall confirmation.} The prospective class-mean study uses $1{,}280$
fit, $1{,}280$ calibration, and $1{,}280$ confirmation entities, balanced across
four types, for $16{,}320$ relation rows. The primary interval uses $10{,}000$
draws that independently resample fit and confirmation entities within type and
refit the complete nested model. The permutation control uses $200$
type-preserving permutations, whose absolute median and $95$th-percentile log-loss
improvements are $0.000076$ and $0.000430$, with a $95$th-percentile Brier improvement of
$0.000169$; adding the reference-answer likelihood instead improves log loss by
$+0.0336$ and Brier score by $+0.0096$. The design matrices of $M_0$ and $M_1$ have
$26$ and $27$ columns, ranks $23$ and $24$, and condition numbers over nonzero
singular values of $14.66$ and $14.99$. A synthetic population with the same design
and a planned improvement of about $0.010$ in log loss and $0.0042$ in Brier score
returns $+0.0106$ ($[0.0064,0.0146]$) and $+0.0044$ ($[0.0026,0.0061]$) over
$2{,}000$ joint-refit draws, so the design detects an effect of the observed size.
The stricter seven-condition decision rule of an earlier protocol, reported as a
secondary analysis, is also met: both interval lower bounds are positive, both
Holm-adjusted tests reject, all four entity types improve on both scores, the
improvements among high-exposure entities are positive ($+0.0193$ in log loss and
$+0.0065$ in Brier score) with every type represented, and the upper bound on
calibration error is at most $0.10$ for every type and probability bin. The cohort, model revisions, direction, prompts,
scorer, baseline, controls, decision rule, and code were fixed before any model
measurement.

\section{Qwen-Scope Replication Details}
\label{app:qwen}

\paragraph{Readout and collection.}
The model is \path{Qwen/Qwen3-8B} at revision \texttt{b968826d}; the SAE is
\path{Qwen/SAE-Res-Qwen3-8B-Base-W64K-L0_100} at revision \texttt{f308c1df}.
The SAE was trained on the Qwen3-8B base model, and we apply its encoder
unchanged to the post-trained checkpoint. Layer $23$, feature $58682$, and
negative orientation are fixed across both Qwen3 recall cohorts. The feature is
read from the prefill residual stream at the final token of the last entity
mention in each bare relation prompt, so its value can differ across relations
of the same entity. Generation is greedy in bfloat16 with a $64$-token limit,
and recall labels come from the deterministic reference matcher.

\paragraph{Two cohorts and the amended permutation limit.}
On the first Qwen3 cohort, the readout improved log loss by $+0.008487$ and
Brier score by $+0.002982$, but the absolute median of the permuted-label
improvements, $0.001554$, exceeded the pre-specified limit of $0.001$, so we do
not count this cohort as a confirmation. For a second cohort, whose entities
share no QID or normalized alias with any earlier Qwen3 cohort, we raised the
median limit to $0.002$ and kept the $95$th-percentile limit at $0.005$. We
made this change after seeing the first result and before any measurement on
the second cohort; file ordering documents this sequence, and the amended protocol was not registered externally.

\paragraph{Controls and uncertainty.}
Under the amended limits, $200$ fit-label permutations that preserve type and
relation give absolute median and $95$th-percentile improvements of $0.001465$
and $0.002522$ for log loss and $0.000466$ and $0.000795$ for Brier score; the log-loss median exceeds the original limit of $0.001$ and meets only the amended limit of $0.002$.
Adding the reference-answer likelihood instead improves log loss by $+0.035163$
and Brier score by $+0.010103$. Class support and design-matrix conditioning
meet their pre-specified checks, the controls were computed before the
confirmation estimates, and intervals use $10{,}000$ joint-refit
entity-bootstrap draws. The pre-specified outcome is positive.

\paragraph{Name collisions.}
The second cohort contains $3{,}840$ unique QIDs and $3{,}838$ unique
normalized names. Two names, \emph{Shelter} and \emph{The Little Things}, refer
to movies in the calibration set and to songs in the confirmation set; neither
appears in the fit set. The collisions affect $12$ calibration and $8$
confirmation relation rows, which the reported result retains.

\section{Llama Replication Details}
\label{app:llama}

We use \path{NousResearch/Meta-Llama-3.1-8B-Instruct} at revision
\path{d10aef7999a2b5ba950ab3974312feeedbfe0b77}. The entities and their fit,
calibration, and confirmation roles match Gemma: $1{,}280$ entities per role,
balanced across four types. The direction is fitted separately on $112$
entities, $14$ for each combination of type and extreme recall label. These
labels come from the upstream Llama base-model recall records, and no cohort
entity name is used in fitting. Leave-one-type-out AUROC over the residual
stream before layers $1$--$31$ selects layer $15$ (macro AUROC $0.934949$),
breaking ties toward earlier layers. The direction is fixed before cohort
collection.

The prompt is \texttt{The \{entity\_type\} '\{entity\}'}, and the score is
read at the final entity token. Recall uses the original bare relation prompts,
greedy decoding, and a $64$-token cap. The baseline includes measured exposure,
missingness, type, relation, name length, and name likelihood from the same
Llama Instruct checkpoint. The L2 nested learners and the $10{,}000$-draw
joint-refit entity bootstrap match the Gemma comparison. Recall is scored by
reference matching, separate from the selection study's stricter criterion.

Across $200$ label permutations, the absolute $95$th-percentile improvements are
$0.000836$ for log loss and $0.000252$ for Brier score. Adding the
reference-answer likelihood instead improves them by $+0.086208$ and
$+0.029394$. The design matrices of $M_0$ and $M_1$ have $26$ and $27$
columns, ranks $23$ and $24$, and condition numbers over nonzero singular
values of $14.57$ and $14.69$. Both primary intervals exceed zero, and every
pre-specified control is met.

\section{Scoring and Selection Reproducibility}
\label{app:scoring}

\paragraph{Reference matching.}
Reference matching reimplements the matcher of \citet{ferrando2025entity}.
Name-valued relations use a token-set similarity above $90$, which can accept
a gold-name mention inside a negated or otherwise incorrect statement.
Numerical tolerances are $\pm5$ years for player birth year, $\pm2$ for movie
release year, $\pm10\%$ for population, $\pm20$ for elevation, and $\pm3$
minutes for duration. Song publication year has zero tolerance. The geographic
matcher takes absolute integer coordinates and accepts a response once two of its
numbers fall within one degree of them, without one-to-one assignment to latitude
and longitude.

\paragraph{Stricter criterion.}
The stricter criterion labels every response alongside reference matching. Name
matches require a contiguous token sequence with no negation cue among the three
preceding tokens of the same sentence, and a one-token reference element
immediately followed by another capitalized name word is not credited. Responses
with two or more line-initial option markers such as \texttt{A.}, \texttt{A)}, or
\texttt{1.}, or with a run of five or more capitalized name tokens, are rejected as
option lists, even when they end by choosing an answer. For multi-valued relations, one
matching gold element suffices. Numerical fields use equality after parsing and
rounding to integers; coordinates require directional latitude and longitude
within $0.1$ degrees. The rule does not cover every way of expressing a valid
answer: aliases are incomplete, local negation is heuristic, and exact
population or duration checks can reject plausible rounding or alternate
references. A matched mention can also sit inside an answer that is otherwise wrong. The rule
is the primary criterion of the prospective selection study; Appendix~\ref{app:human-audit} compares it with the
two-author audit, and Appendix~\ref{app:semantic} with semantic and human labels.
Movie genres are excluded from the eligible relation sets because the stricter
criterion does not score them.

\paragraph{Prospective selection cohort.}
An internal protocol, fixed before generation and not registered externally, specifies the relation set, sampling seed, generation settings, correctness rules,
verification prompt, selection rules, and primary comparisons. Gemma, Gemma~2 9B-IT, and Gemma~2 27B-IT sample $16$ responses per question at $T=1$ with top-$k=50$, no nucleus truncation, and a $24$-token cap (Section~\ref{sec:setup}); Llama, all five Qwen models, and the fresh runs of Appendix~\ref{app:measured-runtime} use $T=1$, $top\_p=1$, $top\_k=0$, and the same cap. The format filter discards empty outputs and outputs
with at least two enumerated-option markers (parenthesized letters, numbered
items, or bullets) or at least six asterisks. It retains $7{,}214$ of the $9{,}600$ Gemma samples
($75\%$), against $9{,}363$ in Llama ($97.5\%$) and $9{,}443$ in Qwen3 ($98.4\%$).
Requiring at least eight finite-scored candidates leaves $541$ Gemma questions over
$451$ entities and $6{,}864$ candidates, a mean of $12.7$ per question. Under the stricter criterion, $204$ questions are covered and $199$ are mixed. The
main selection tables use this population throughout. Candidate data, scores,
and both labels will be released with the code.

\paragraph{Development sampling cohort and earlier same-question sample.} The development
sampling cohort (Table~\ref{tab:cohorts}) is sampled at $T=0.3$, $0.7$, $1.0$, and $1.3$.
Its temperature comparisons use the questions eligible at all four temperatures,
$120$ name-valued and $170$ answerable questions; the answerable questions exclude
population questions because the prompt does not specify the date or geographic scope
needed to interpret the reference integer. At $T=1$, its $152$ name-valued questions give the development selection comparison in Appendix~\ref{app:selection-tables}. The
earlier same-question sample gives the $245$-question analysis split by greedy outcome
(Table~\ref{tab:query-discrimination}), with its original reference-matching labels.

\paragraph{Scoring, selection, and uncertainty.}
Likelihood is teacher-forced on decoded answer text under the untempered
model, so it does not depend on the top-$k$ truncation used for sampling, and
padding is excluded from score averages.
The verification prompt supplies the question and proposed answer but no gold
answer. Its Yes/No normalization defines a $P(\mathrm{True})$-style ranking score.
Voting uses the normalized answer keys of Section~\ref{sec:setup}.
Appendix~\ref{app:weighted-voting} evaluates CISC's confidence-softmax
aggregation on the prospective cohort.

At $K=16$, each eligible pool contributes all its retained
candidates. Paired intervals in the selection studies, including the conversion interval, use $20{,}000$ entity-bootstrap draws with seed $20260923$. Budget and temperature curves use $400$ candidate subsamples per question. Entity-clustered intervals preserve dependence among questions and responses but condition on the saved candidates and labels. Differences are computed from unrounded estimates and may differ slightly from differences between displayed values.

\section{Sparse Autoencoder Latents}
\label{app:sae}

On the prospective Gemma recall cohort, we also evaluate the two SAE latents published by
\citet{ferrando2025entity}. Because they were selected on a population whose extreme
labels had already been examined, their intervals hold the fitted models fixed.
Both improve recall prediction, less than the class-mean readout does (Appendix~\ref{app:recall-tables},
Table~\ref{tab:construct-sparse}).

\paragraph{Discovery.} Searching the $25$ Gemma Scope residual-stream SAEs of width $16$k for layers $0$--$24$ \citep{lieberum2024gemmascope} over $24{,}896$ known or unknown assignments from a
catalog of $34{,}734$ entities, with the $<2\%$ generality filter on $10{,}000$ Pile
rows \citep{gao2020pile}, recovers the two latents selected by
\citet{ferrando2025entity} (Table~\ref{tab:discovery}). Model and SAE files are fixed
by Hugging Face revision and file checksum.

\begin{table}[H]
\centering
\small
\setlength{\tabcolsep}{5pt}
\caption{\textbf{Discovery recovers both published latents.} Scores are bfloat16 on
CUDA; float32 scores on Apple MPS differ slightly.}
\label{tab:discovery}
\begin{tabular}{llrrr}
\toprule
Direction & SAE & Latent & MaxMin & Pile frequency \\
\midrule
Known & \texttt{layer\_12/width\_16k/average\_l0\_82} & $7{,}957$ & $0.64049$ & $0.0150$ \\
Unknown & \texttt{layer\_14/width\_16k/average\_l0\_84} & $11{,}898$ & $0.27795$ & $0.0076$ \\
\bottomrule
\end{tabular}
\end{table}

\paragraph{Refusal steering.} We reproduce the refusal intervention of
\citet{ferrando2025entity} on $400$ unknown-entity prompts, $100$ per type, under $24$
conditions: unsteered, a zero direction, the two target latents, and ten matched
random latents per side. Generation is greedy with a $30$-token limit, and the
decoder direction is added with coefficient $400$ from the entity-final token through
the end of the prompt, giving $9{,}600$ generations. Refusals are counted with the
automatic matcher of the reference implementation (Table~\ref{tab:behavior}); the
original study's $320$-row annotation packet has no adjudicated labels. Averaging
prompt-level refusal differences within each of the $351$ type--entity groups ($82$
player, $90$ city, $99$ movie, and $80$ song), then within type, and then across
types, latent $11{,}898$ raises refusal by $0.6204$ ($[0.5757,0.6656]$), with
positive effects in every type (player $0.6667$, city $0.3037$, movie $0.7071$, song
$0.8042$); the ten matched random latents reach at most $0.1574$. Zero-direction
generations are token-identical to unsteered ones on all $400$ prompts, the steering hooks modify no generated-token position, the maximum steering error is below $10^{-5}$, and
an independent rerun reproduces every token sequence and label. In a post hoc analysis of the same generations, the known latent $7{,}957$ lowers refusal by $0.1802$ ($[0.1293,0.2300]$), but the strongest matched random latent lowers it by $0.1764$ (paired contrast $[-0.0429,0.0505]$), and five of the ten target-minus-random intervals include zero. Unlike the unknown-side effect, the known-side effect is therefore not specific to the latent, and this reproduction does not support the known-side result of \citet{ferrando2025entity}. The steering adds
a decoder direction to the residual stream at prompt positions only;
\citet{ferrando2025entity} report coefficients of about $400$--$550$ and also modify
all model weights by orthogonalization, which we do not reproduce.

\begin{table}[H]
\centering
\small
\setlength{\tabcolsep}{5pt}
\caption{\textbf{Refusal steering on Gemma~2 2B-IT.} Refusals among $400$
unknown-entity prompts, counted by the automatic matcher of the reference
implementation.}
\label{tab:behavior}
\begin{tabular}{lr}
\toprule
Condition & Refusals \\
\midrule
Unsteered & $128$ \\
Zero direction & $128$ \\
Unknown latent $11{,}898$ & $374$ \\
Known latent $7{,}957$ & $55$ \\
Strongest matched random latent, unknown side & $194$ \\
\bottomrule
\end{tabular}
\end{table}

\paragraph{Error ranking and selective answering.} The $512$-entity chat cohort (Table~\ref{tab:cohorts}) yields $2{,}176$ chat responses with human labels relative to the reference
answer: $476$ correct answers, $1{,}204$ wrong attempted answers, $412$ refusals or
non-answers, and $84$ undeterminable responses. Four entity-disjoint rotations fit,
calibrate, and evaluate a logistic baseline on exposure, missingness, type, and relation, alone and with the standardized latent-$11{,}898$ activation added. Among
attempted answers, the latent alone ranks wrong above correct answers with AUROC
$0.6598$ ($[0.6292,0.6880]$), against $0.4959$ for a random ordering; the baseline reaches $0.8804$ ($[0.8589,0.9002]$), and $0.8819$ ($[0.8609,0.9013]$) with the latent added
(Table~\ref{tab:detector}). Among all determinate responses, the latent's AUROC for
wrong assertions against correct answers and refusals is $0.5366$
($[0.5076,0.5651]$), and its AUROC for refusals and non-answers is $0.6547$. At about
$80\%$ retention of attempted answers, the latent keeps $441$ of $476$ correct
answers ($92.6\%$) and withholds $296$ of $1{,}204$ wrong ones ($24.6\%$); the baseline keeps $469$ ($98.5\%$) and withholds $335$ ($27.8\%$) (Table~\ref{tab:selective}). The
bootstrap envelope for the latent's risk difference from random ordering is
$[-10.83,+2.30]$ points, and for adding the latent to the baseline it is $[-6.58,+6.39]$. Relations differ
sharply: among the same $128$ city entities, country yields $76$ correct answers,
elevation one, and location and population none; $38$ of the $84$ undeterminable
responses concern movie genres.

\begin{table}[H]
\centering
\small
\setlength{\tabcolsep}{5pt}
\caption{\textbf{Ranking wrong against correct attempted answers.} $512$-entity chat cohort; AUROCs compare responses within each evaluation fold, with $2{,}000$ type- and
fold-stratified entity-bootstrap draws.}
\label{tab:detector}
\begin{tabular}{lcc}
\toprule
Predictor & AUROC & 95\% interval \\
\midrule
Latent $11{,}898$ & $0.6598$ & $[0.6292,0.6880]$ \\
Random ordering & $0.4959$ & $[0.4577,0.5344]$ \\
Baseline: exposure, missingness, type, relation & $0.8804$ & $[0.8589,0.9002]$ \\
Baseline and latent $11{,}898$ & $0.8819$ & $[0.8609,0.9013]$ \\
\bottomrule
\end{tabular}
\end{table}

\begin{table}[H]
\centering
\small
\setlength{\tabcolsep}{4pt}
\caption{\textbf{Selective answering at about $80\%$ retention.} Thresholds are
selected on calibration folds; counts use the human labels, and the risk bounds let
every released undeterminable response be either correct or wrong.}
\label{tab:selective}
\begin{tabular}{lrrrrl}
\toprule
Policy & Retention & Correct & Wrong & Undeterminable & Risk bounds \\
\midrule
Answer all & $100\%$ & $476$ & $1{,}204$ & $84$ & $[68.25,73.02]\%$ \\
Random & $78.57\%$ & $372$ & $948$ & $65$ & $[68.45,73.14]\%$ \\
Latent $11{,}898$ & $80.30\%$ & $441$ & $908$ & $64$ & $[64.26,68.79]\%$ \\
Baseline & $79.64\%$ & $469$ & $869$ & $72$ & $[61.63,66.74]\%$ \\
Baseline and latent & $80.00\%$ & $470$ & $874$ & $74$ & $[61.64,66.85]\%$ \\
\bottomrule
\end{tabular}
\end{table}

\paragraph{Chat and bare-format answers.} The same $2{,}176$ entity--relation rows also
have greedy bare-format outcomes. The chat model answers $602$ of $634$ bare-format
hits ($95.0\%$) and $1{,}078$ of $1{,}458$ bare-format misses ($73.9\%$); of the
answered misses, $1{,}015$ are wrong and $63$ correct, and $32$ bare-format hits become
refusals or non-answers (Table~\ref{tab:format-routing}).

\begin{table}[H]
\centering
\small
\setlength{\tabcolsep}{5pt}
\caption{\textbf{Chat outcomes by bare-format outcome.} $512$-entity chat cohort;
bare-format hits and misses use reference matching and chat outcomes the human
labels.}
\label{tab:format-routing}
\begin{tabular}{lrrrr}
\toprule
Bare format & Chat correct & Chat wrong & Chat refusal & Undeterminable \\
\midrule
Hit & $413$ & $189$ & $32$ & $33$ \\
Miss & $63$ & $1{,}015$ & $380$ & $51$ \\
\bottomrule
\end{tabular}
\end{table}

\paragraph{Pretrained and instruction-tuned models.} The $512$-entity chat cohort comes from a paired study of the pretrained and instruction-tuned Gemma~2 2B models. In the
pretrained model, latent $11{,}898$ improves recall prediction by $0.000564$ in Brier
score and $0.001660$ in log loss, with both intervals above zero; in the
instruction-tuned model both intervals include zero, and the difference between the
two models is positive.

\section{Confidence-Weighted Voting}
\label{app:weighted-voting}

We apply CISC's confidence-softmax voting rule
\citep{taubenfeld2025confidence} to the same $541$ prospective questions.
Candidate weights are proportional to $\exp(c_i/\tau)$, where $\tau$ is the
aggregation temperature and $c_i=\exp(s_{{\rm LP},i})$ or $s_{{\rm True},i}$. We sum weights over the
normalized answer strings of Section~\ref{sec:setup} and select the highest-weight group.
Within tied groups, the highest-confidence candidate is returned; exact score
ties average correctness. We report the full fixed grid
$\tau\in\{0.05,0.1,0.3,1\}$ without choosing a temperature from outcomes; this comparison is secondary to the prospectively specified plurality test. Intervals are computed as in Appendix~\ref{app:scoring}.

\begin{table}[H]
\centering
\small
\setlength{\tabcolsep}{5pt}
\caption{\textbf{Verification improves confidence-weighted selection.}
Accuracies are percentages; differences and paired $95\%$ intervals are
percentage points. Scores, candidates, and aggregation temperatures are identical between criteria. Intervals are pointwise.}
\label{tab:weighted-voting}
\begin{tabular}{llrrrl}
\toprule
Criterion & $\tau$ & Likelihood & $P(\mathrm{True})$ & Difference & 95\% CI \\
\midrule
Stricter & $0.05$ & $21.63$ & $27.54$ & $+5.91$ & $[3.33,8.58]$ \\
 & $0.1$ & $22.00$ & $27.54$ & $+5.55$ & $[2.99,8.16]$ \\
 & $0.3$ & $22.74$ & $27.36$ & $+4.62$ & $[2.18,7.16]$ \\
 & $1$ & $23.11$ & $27.54$ & $+4.44$ & $[2.19,6.80]$ \\
\midrule
Reference & $0.05$ & $26.80$ & $27.91$ & $+1.11$ & $[-1.13,3.42]$ \\
 & $0.1$ & $26.99$ & $27.91$ & $+0.92$ & $[-1.30,3.19]$ \\
 & $0.3$ & $26.99$ & $27.73$ & $+0.74$ & $[-1.51,3.01]$ \\
 & $1$ & $26.99$ & $27.91$ & $+0.92$ & $[-1.25,3.09]$ \\
\bottomrule
\end{tabular}
\end{table}

In Llama, under the stricter criterion, the same four aggregation temperatures give
gains of $1.34$ points
at $\tau=0.05$ and $0.1$ and $1.51$ points at $\tau=0.3$ and $1$, close to the
$1.67$-point primary plurality gain (Table~\ref{tab:llama-selection}), with
paired intervals $[-0.38,3.08]$, $[-0.34,3.04]$, $[-0.18,3.22]$, and
$[-0.14,3.25]$ that include zero. Reference-matching gains are $1.17$--$1.51$
points, with intervals that also include zero.

\section{Blinded Human Annotation Audit}
\label{app:human-audit}

\subsection{Annotation protocol}
Two authors independently label all $400$ candidates in the audit
set. Each sees only the factual question, the reference answer, and the
candidate answer, and is blind to automatic labels, model scores, selector
identities, and the other annotator's labels. Labels are \emph{correct},
\emph{incorrect}, or \emph{ambiguous} under the factual meaning of the question;
legitimate multi-valued answers are not rejected merely for being lists.
Ambiguous labels are retained as a third category.

\subsection{Sampling design}
The audit set combines both argmax selections from $100$ uniformly sampled
eligible questions, selected disagreements between the reference and stricter
criteria, random candidates, and enriched agreement controls. The uniform
component gives a representative sample of selected outputs; the enriched
components characterize the two criteria on informative cases.

\subsection{Agreement}
Three-way agreement between the two annotators is $84.25\%$ (Cohen's
$\kappa=0.729$).

\subsection{Agreement with the automatic criteria}
Table~\ref{tab:human-audit-endpoint} cross-tabulates the annotators' labels
against the automatic criteria; in this set every stricter acceptance is also
a reference acceptance. Among the $91$ responses both rules accept, the
annotators label only $2$ and $3$ incorrect. Of the $46$ responses accepted only by reference matching, the annotators label $27$ and $30$ incorrect and $17$ and $15$
correct; the correct ones are mostly multiple-choice outputs that end by
choosing the right option, which the stricter criterion rejects as enumerations.

\begin{table}[H]
\centering
\small
\setlength{\tabcolsep}{5pt}
\caption{\textbf{Annotator labels by automatic outcome.} Correct / incorrect /
ambiguous counts over the enriched audit set; these are not population
rates.}
\label{tab:human-audit-endpoint}
\begin{tabular}{lrcc}
\toprule
Automatic outcome & Responses & Annotator A & Annotator B \\
\midrule
Both criteria accept & $91$ & $79/2/10$ & $69/3/19$ \\
Reference matching only & $46$ & $17/27/2$ & $15/30/1$ \\
Neither criterion & $263$ & $1/180/82$ & $3/219/41$ \\
\bottomrule
\end{tabular}
\end{table}

\subsection{Disputed selections}
All $31$ argmax selections accepted only by reference matching, $28$ by
likelihood and $3$ by verification, are in the audit set. Annotator A labels
likelihood's $28$ as $10$ correct, $17$ incorrect, and $1$ ambiguous, and
Annotator B as $9$ correct and $19$ incorrect; both label verification's $3$
correct or ambiguous. Replacing the stricter label with each annotator's
label on these $31$ selections, counting ambiguous as incorrect, gives
likelihood and verification argmax accuracies of $23.29\%$ and $27.17\%$
(Annotator A) and $23.11\%$ and $27.36\%$ (Annotator B): advantages of $3.88$
and $4.25$ points, against $5.36$ under the stricter criterion and $0.74$ under reference matching.

\subsection{Selector comparison on uniformly sampled questions}
Table~\ref{tab:human-audit-selector} reports correct argmax selections on the
$100$ uniformly sampled questions. The annotator comparisons are $25$ versus
$28$ and $24$ versus $23$, with paired intervals of $[-4.9,10.9]$ and
$[-9.0,7.1]$ points; the automatic rules give $22$ versus $30$ (stricter) and
$28$ versus $31$ (reference). With $14\%$ discordant questions under the stricter
criterion, a $100$-question paired comparison has about $27\%$ power to detect a
$5$-point difference, so Appendix~\ref{app:semantic} gives the human selection gain
on all $541$ questions.

\begin{table}[H]
\centering
\small
\setlength{\tabcolsep}{5pt}
\caption{\textbf{Correct argmax selections on the $100$ uniformly sampled
questions.}}
\label{tab:human-audit-selector}
\begin{tabular}{lrrrr}
\toprule
Selector & Annotator A & Annotator B & Stricter criterion & Reference matching \\
\midrule
Likelihood & $25$ & $24$ & $22$ & $28$ \\
$P(\mathrm{True})$ & $28$ & $23$ & $30$ & $31$ \\
\bottomrule
\end{tabular}
\end{table}

\section{Llama Ranking and Selection}
\label{app:llama-selection}

The Llama selection study uses the same model revision as its recall study
and the same $600$ questions as the prospective Gemma selection cohort.
New responses are sampled in bfloat16 on an RTX A5000 with the settings of Appendix~\ref{app:scoring}, and the format filter is applied to all $16$ saved samples. Likelihood is recomputed by
teacher forcing, and verification uses the model's chat template without a
reference answer. The eligibility rule retains $598$ questions,
$489$ entity clusters, and $9{,}349$ candidates. Llama's original-prompt
comparison is specified before outcome analysis;
Appendix~\ref{app:prompt-robustness} reports the three additional wordings.

\begin{figure}[H]
\centering
\includegraphics[width=\linewidth]{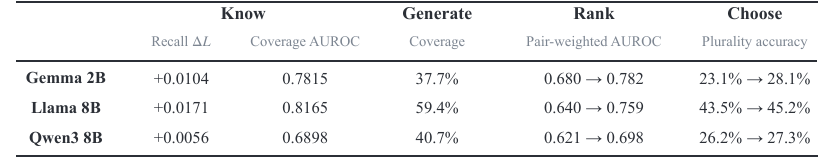}
\caption{\textbf{Factual competence profiles.} Recall improvement, the readout's
AUROC for $K=16$ coverage (Section~\ref{sec:know-choose}; Gemma on the $521$
questions used there), $K=16$ coverage on each model's eligible questions
($541$, $598$, and $600$), pair-weighted AUROC, and plurality accuracy.
Arrows compare mean log-likelihood with $P(\mathrm{True})$.}
\label{fig:profiles}
\end{figure}

Under the stricter criterion, $355$ questions ($59.36\%$) are covered, and plurality
conversion rises from $73.24\%$ with likelihood to $76.06\%$ with verification,
which gains $17$ correct selections and loses $7$.
Pair-weighted AUROC increases from $0.6401$ to $0.7586$, a gain of $0.1185$
($[0.084,0.154]$; Figure~\ref{fig:profiles}),
across $302$ mixed questions and $10{,}698$ pairs. Query-macro AUROC increases from $0.6341$
to $0.7593$, a paired gain of $0.1252$ with interval $[0.0956,0.1548]$.
Exact score ties use uniform selection weight, so accuracy averages their
labels. Likelihood argmax and likelihood plurality select the same answer group
on $588$ of the $598$ questions and never differ in correctness under either
automatic rule, so their accuracies coincide in Table~\ref{tab:llama-selection}. The complete
selection contrasts are in Table~\ref{tab:llama-selection}.

The $1.67$-point plurality gain has an interval above
zero in this comparison and on the filtered pools of Appendix~\ref{app:rank-analysis};
under the other three prompt wordings, confidence-weighted voting, fresh samples,
matched compute, and semantic labels ($1.51$ points, $[-0.10,3.15]$), the interval
includes zero (Appendices~\ref{app:prompt-robustness}, \ref{app:weighted-voting},
\ref{app:measured-runtime}, \ref{app:compute-accounting}, and~\ref{app:semantic}).

\begin{table}[H]
\centering
\small
\setlength{\tabcolsep}{5pt}
\caption{\textbf{Llama selection on matched input questions.} Original
verification prompt, $598$ eligible questions. Accuracies are percentages;
differences and paired entity-clustered $95\%$ intervals are percentage points.}
\label{tab:llama-selection}
\begin{tabular}{llrrrl}
\toprule
Criterion & Rule & Likelihood & $P(\mathrm{True})$ & Difference & 95\% CI \\
\midrule
Stricter & Argmax & $43.48$ & $44.82$ & $+1.34$ & $[-0.48,3.16]$ \\
 & Plurality & $43.48$ & $45.15$ & $+1.67$ & $[0.05,3.30]$ \\
Reference & Argmax & $44.98$ & $46.15$ & $+1.17$ & $[-0.61,2.96]$ \\
 & Plurality & $44.98$ & $46.49$ & $+1.51$ & $[-0.16,3.18]$ \\
\bottomrule
\end{tabular}
\end{table}

Restricting to the $540$ questions eligible for both Gemma and Llama preserves the contrast: Llama's pair-weighted AUROC rises from $0.6386$ to $0.7583$,
while plurality accuracy rises from $43.89\%$ to $45.56\%$, a $1.67$-point gain
($[-0.06,3.41]$).
Each model generates and filters its own candidates, so the pools differ even
on shared questions.

\section{Rank Transitions and the Text Filter}
\label{app:rank-analysis}

\begin{figure}[H]
\centering
\captionsetup[subfigure]{font=small,labelfont=bf,justification=raggedright,
singlelinecheck=false,skip=4pt}
\begin{subfigure}[t]{0.40\linewidth}
\centering
\includegraphics[width=\linewidth]{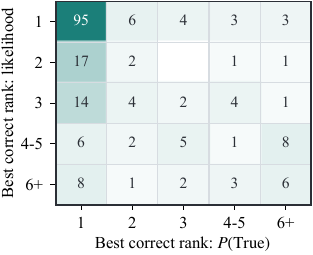}
\caption{\textbf{Gemma.} A correct answer reaches first rank on $45$ questions and leaves it on $16$.}
\end{subfigure}\hspace{0.08\linewidth}
\begin{subfigure}[t]{0.40\linewidth}
\centering
\includegraphics[width=\linewidth]{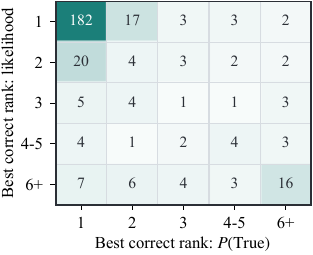}
\caption{\textbf{Llama.} A correct answer reaches first rank on $36$ questions and leaves it on $25$.}
\end{subfigure}
\caption{\textbf{Rank transitions.} Stricter criterion. Mixed questions counted by the rank of the highest-scoring correct candidate under likelihood (rows) and verification (columns), under direct score
ordering.}
\label{fig:rank-transitions}
\end{figure}

For each question with both correct and incorrect candidates, we record the rank of
its highest-scoring correct candidate under score $s$: one plus the number of
candidates scoring strictly above it. Figure~\ref{fig:rank-transitions} counts
transitions of this rank between likelihood and verification. Verification
places a correct answer at the highest score on $140$ of Gemma's $199$ mixed
questions ($70.4\%$) and $218$ of Llama's $302$ ($72.2\%$); likelihood does so
on $111$ ($55.8\%$) and $207$ ($68.5\%$). This rank improves on
$65$ Gemma and $67$ Llama questions and worsens on $34$ and $44$. Because an
incorrect candidate may share the maximum score, the numbers of questions whose correct answer
reaches or leaves first rank in Figure~\ref{fig:rank-transitions} need not equal the change in tie-averaged
selection accuracy; plurality also depends on normalized-answer vote counts
(Section~\ref{sec:verification}).

The text filter removes explicit option lists, negated
responses, and refusal language using the answer text alone. It imposes no length
or single-sentence requirement. Both scorers are
evaluated on the same retained candidates for each question. Table~\ref{tab:format-control}
shows that verification continues to improve within-question factual ranking
after these outputs are removed.

\paragraph{Option lists and the text filter.} Gemma generates many option lists
at $T=1$ (Appendix~\ref{app:scoring}), and mean log-likelihood favors them: its
argmax is an option list on $61$ of Gemma's $541$ questions, against $17$ for summed
log-likelihood and $11$ for verification; in Llama the counts are $11$, $11$, and
$7$. Verification avoids most option lists without the text filter, and
rejecting them is usually correct: the two-author audit labels most of
likelihood's disputed option-list selections incorrect
(Appendix~\ref{app:human-audit}). Applying the text filter before likelihood
raises Gemma plurality accuracy to $25.69\%$, still below the $28.10\%$ of
unfiltered verification (Table~\ref{tab:extra-baselines}). When both selectors use text-filtered pools, on the $540$ Gemma questions that keep at least two candidates, verification still raises Gemma plurality
accuracy by $3.52$ points ($[1.57,5.47]$), most of its full $4.99$-point gain;
under semantic labels, it keeps $3.15$ points ($[1.2,5.1]$) of $3.88$. Pair-weighted
AUROC still rises from $0.722$ to $0.776$ in Gemma and from $0.633$ to $0.754$ in
Llama, and the Llama plurality gain on the filtered pools is $1.51$ points
($[0.02,3.05]$), close to its full-pool gain (Tables~\ref{tab:format-control}
and~\ref{tab:format-selection}).

\begin{table}[H]
\centering
\small
\setlength{\tabcolsep}{5pt}
\caption{\textbf{Factual ranking after the text filter.} Stricter criterion. Pair-weighted AUROC
uses mixed questions; the gain and interval average question-level differences
with entity-clustered resampling.}
\label{tab:format-control}
\begin{tabular}{lrrrrl}
\toprule
Model & Candidates & Mixed questions & Likelihood & $P(\mathrm{True})$ & Query-macro gain [95\% CI] \\
\midrule
Gemma & $5{,}649$ & $184$ & $0.722$ & $0.776$ & $+0.0667$ $[0.0257,0.1072]$ \\
Llama & $8{,}242$ & $287$ & $0.633$ & $0.754$ & $+0.1234$ $[0.0942,0.1521]$ \\
\bottomrule
\end{tabular}
\end{table}

After filtering, $540$ Gemma and $598$ Llama questions retain at least two
candidates. Table~\ref{tab:format-selection} reports selection on those same
filtered pools; the full-pool prospective results remain in Table~\ref{tab:ptrue}.
On the filtered Llama pools, verification plurality gains $15$ correct
selections and loses $6$.

\begin{table}[H]
\centering
\small
\setlength{\tabcolsep}{5pt}
\caption{\textbf{Selection after the text filter.} Stricter criterion. Accuracies are percentages;
gains and entity-clustered $95\%$ intervals are percentage points.}
\label{tab:format-selection}
\begin{tabular}{llrrl}
\toprule
Model & Rule & Likelihood & $P(\mathrm{True})$ & Gain [95\% CI] \\
\midrule
Gemma & Argmax & $25.37$ & $27.78$ & $+2.41$ $[0.22,5.12]$ \\
 & Plurality & $25.74$ & $29.26$ & $+3.52$ $[1.57,5.47]$ \\
Llama & Argmax & $44.15$ & $45.48$ & $+1.34$ $[-0.43,3.10]$ \\
 & Plurality & $44.15$ & $45.65$ & $+1.51$ $[0.02,3.05]$ \\
\bottomrule
\end{tabular}
\end{table}

\paragraph{Relation groups.} Table~\ref{tab:relation-groups} splits both models into high-prior and low-prior
relations: the four relations where proposing the $16$ most common answers of the
relation covers at least $10\%$ of questions (country, teams, genres, and
publication year; Appendix~\ref{app:decoding}), and the other six, where it covers
$2.9\%$. In Gemma, the verification advantage in ranking and selection comes
mainly from the high-prior relations, which account for $23$ of the $27$ net
additional correct selections under the stricter criterion; in the low-prior
relations, the two scores
rank comparably and the selection gain is $1.28$ points. In Llama, the ranking
advantage holds in both groups, and the selection gain again comes from the
high-prior relations. The difference between the Llama gains has the same sign as
in Gemma, and its interval includes zero.

\begin{table}[H]
\centering
\small
\setlength{\tabcolsep}{4pt}
\caption{\textbf{Ranking and selection by relation group.} High-prior relations
are the four with at least $10\%$ common-answer coverage; $n$
counts questions. Stricter criterion; pair-weighted AUROC over mixed questions;
plurality accuracy in percent; gains and entity-clustered $95\%$ intervals in
points. High $-$ low rows give the difference between the two gains.}
\label{tab:relation-groups}
\setlength{\tabcolsep}{3pt}
\begin{tabular}{llrrcrrl}
\toprule
Model & Relations & $n$ & Mixed & AUROC & Likelihood & $P(\mathrm{True})$ & Gain [95\% CI] \\
\midrule
Gemma & High-prior & $229$ & $113$ & $0.618\rightarrow0.810$ & $27.51$ & $37.55$ & $+10.04$ $[5.53,14.31]$ \\
 & Low-prior & $312$ & $86$ & $0.744\rightarrow0.753$ & $19.87$ & $21.15$ & $+1.28$ $[-0.98,3.59]$ \\
 & High $-$ low & --- & --- & --- & --- & --- & $+8.76$ $[3.5,13.7]$ \\
\midrule
Llama & High-prior & $283$ & $162$ & $0.631\rightarrow0.756$ & $47.70$ & $51.59$ & $+3.89$ $[0.34,7.46]$ \\
 & Low-prior & $315$ & $140$ & $0.649\rightarrow0.761$ & $39.68$ & $39.37$ & $-0.32$ $[-3.73,3.08]$ \\
 & High $-$ low & --- & --- & --- & --- & --- & $+4.20$ $[-0.75,9.15]$ \\
\bottomrule
\end{tabular}
\end{table}

\section{Verification Prompt Robustness}
\label{app:prompt-robustness}

We re-score the saved Gemma and Llama candidates with the original instruction
and the three rewordings of Appendix~\ref{app:prompts}. The question, answer, chat template, and first-response scoring position
remain fixed. Scores normalize over the named affirmative and negative
tokens; reversing their instruction order does not reverse the positive class.
All variants are reported without selecting a prompt using evaluation labels. Appendix~\ref{app:prompts} gives the full prompt format.

Gemma is rescored with the original checkpoint in float32 and Llama in
bfloat16. For Gemma, the original instruction reproduces every argmax and
plurality accuracy under both criteria and gives pair-weighted stricter AUROC
$0.7827$, against $0.7820$ for the prospective scores in Table~\ref{tab:ptrue}.
The additional wordings, with pointwise intervals, are secondary to the
prospective Gemma test.

All four wordings improve Gemma plurality selection under the stricter
criterion, with gains of $3.33$--$4.99$ points and all paired intervals above
zero (Table~\ref{tab:prompt-robustness}). Argmax gains are
$2.96$--$5.36$ points; the No/Yes interval touches zero, while the
other three exclude it. In Llama, all wordings improve query-macro ranking,
with gains of $0.1106$--$0.1252$ and intervals above zero. Its plurality
gains are $0.84$--$1.67$ points; the original wording's interval
excludes zero, and the other three include it.

\begin{table}[H]
\centering
\small
\setlength{\tabcolsep}{4pt}
\caption{\textbf{Verification under four prompt wordings.} Stricter
criterion; pair-weighted AUROC and selection accuracy for $P(\mathrm{True})$.
The last column is its paired plurality gain over likelihood, in percentage
points with a pointwise $95\%$ entity-clustered interval. Gemma has $541$
eligible questions; Llama has $598$.}
\label{tab:prompt-robustness}
\begin{tabular}{llrrrl}
\toprule
Model & Wording & Pair-weighted AUROC & Argmax & Plurality & Plurality gain [95\% CI] \\
\midrule
Gemma & Original & $0.7827$ & $26.80\%$ & $28.10\%$ & $+4.99\ [2.7,7.3]$ \\
 & Short & $0.7806$ & $26.80\%$ & $27.91\%$ & $+4.81\ [2.60,7.12]$ \\
 & True/False & $0.7697$ & $25.69\%$ & $27.17\%$ & $+4.07\ [1.83,6.46]$ \\
 & No/Yes & $0.7479$ & $24.40\%$ & $26.43\%$ & $+3.33\ [1.09,5.70]$ \\
\midrule
Llama & Original & $0.7586$ & $44.82\%$ & $45.15\%$ & $+1.67\ [0.05,3.30]$ \\
 & Short & $0.7568$ & $44.65\%$ & $44.98\%$ & $+1.51\ [-0.20,3.20]$ \\
 & True/False & $0.7460$ & $44.15\%$ & $44.31\%$ & $+0.84\ [-0.75,2.45]$ \\
 & No/Yes & $0.7555$ & $44.82\%$ & $44.98\%$ & $+1.51\ [-0.45,3.50]$ \\
\bottomrule
\end{tabular}
\end{table}

Reference matching attenuates the Gemma plurality gain for every wording:
$+1.29$ points for the original instruction (interval $[-0.8,3.4]$),
$+1.11$ for the short instruction ($[-0.93,3.18]$), $+0.37$ for True/False
($[-1.69,2.49]$), and $-0.37$ for No/Yes ($[-2.59,1.87]$).
All four intervals include zero. The complete argmax and plurality results
under both criteria will be released with the code.

\paragraph{Paired criterion-by-evaluator interactions.}
For each question we also compute the verification-minus-likelihood gain
under the stricter criterion minus that gain under reference matching, keeping both
selected outputs fixed. In Gemma, across all four wordings, the argmax interactions range from
$4.62$ to $4.81$ points, and every plurality interaction equals $3.70$;
all eight paired intervals exclude zero. In Llama the interactions range
from $-0.17$ to $+0.25$ points, with all intervals including zero.

\section{Token-Accounted Candidate Budgets}
\label{app:compute-accounting}

Verification adds one model evaluation per candidate. For Llama, we evaluate candidate budgets $K\in\{1,2,4,8,16\}$ using
$128$ nested random subsamples per eligible question and the original
verification prompt. The accounting proxy charges each candidate its prompt
and decoded-answer token counts, plus the verification input tokens when
verification is used. Likelihood is assumed to be computed during generation.
The proxy gives no shared-prefill discount and ignores padding, attention
complexity, and batching; Appendix~\ref{app:measured-runtime} reports measured
time.

\begin{table}[H]
\centering
\small
\setlength{\tabcolsep}{3.5pt}
\caption{\textbf{Selection and token-accounted cost.} Llama, $598$ questions,
stricter criterion, plurality with the named tie-break. Accuracy and coverage
are percentages; gain is verification minus likelihood accuracy, in points. Cost is
mean accounted tokens per question.}
\label{tab:token-budgets}
\begin{tabular}{rrrrrrr}
\toprule
$K$ & Coverage & Likelihood accuracy & Verification accuracy & Gain & Likelihood cost & Verification cost \\
\midrule
$1$ & $36.75$ & $36.75$ & $36.75$ & $0.00$ & $52$ & $121$ \\
$2$ & $44.55$ & $38.98$ & $40.91$ & $+1.93$ & $104$ & $241$ \\
$4$ & $50.25$ & $40.50$ & $43.40$ & $+2.90$ & $208$ & $482$ \\
$8$ & $55.04$ & $41.94$ & $44.65$ & $+2.71$ & $416$ & $965$ \\
$16$ & $59.36$ & $43.48$ & $45.15$ & $+1.67$ & $812$ & $1{,}885$ \\
\bottomrule
\end{tabular}
\end{table}

A randomized mixture of verification at $K=4$ and $K=8$, using the latter
with probability $0.684$, matches likelihood's $K=16$ expected proxy cost.
Its accuracy is $44.26\%$ versus $43.48\%$, a $0.78$-point difference with
interval $[-1.14,2.73]$. On an A5000, generation for all $600$ questions
takes $509.3$ seconds, teacher-forced likelihood rescoring adds $124.3$
seconds, and original-prompt verification of the $9{,}363$ retained responses,
before question eligibility is applied, takes $187.1$ seconds, excluding
checkpoint loading.

\section{Decoding and Guessing Baselines}
\label{app:decoding}

\paragraph{Greedy decoding.} On the prospective cohort, greedy decoding with the
same checkpoint, prompts, precision, and $24$-token limit answers $27.54\%$ of
the $541$ eligible questions under the stricter criterion ($30.31\%$ under reference matching), against $17.43\%$ for a single $T=1$ sample. Verification
plurality at $K=16$ is level with the greedy answer ($+0.55$ points,
$[-2.3,3.4]$), whereas summed-likelihood and likelihood plurality fall below it
($-2.96$, $[-5.78,-0.18]$; $-4.44$, $[-7.39,-1.47]$).

\paragraph{Guessing baseline.} Following \citet{yona2025keep}, a prompt-agnostic
baseline proposes the $K$ reference answers most common among other entities'
questions of the same relation, scored by the stricter criterion. It covers $6.3\%$,
$10.0\%$, $15.7\%$, $21.4\%$, and $26.8\%$ of the $541$ questions at
$K=1,2,4,8,16$, against sampled coverage of $17.4\%$ to $37.7\%$
(Table~\ref{tab:selection-budget}). It is strong for a city's country ($76.8\%$
at $K=16$), a player's teams ($52.1\%$), song genres ($50.0\%$), and publication
year ($46.6\%$), and covers $2.9\%$ of the other six relations, where sampled
coverage grows from $13.0\%$ to $28.5\%$.

\paragraph{Temperature.} The development sampling cohort (Appendix~\ref{app:scoring}) holds its $120$ name-valued and $170$ answerable questions fixed across temperatures, using plurality with a likelihood tie-break at $K=16$. In both cohorts, coverage peaks at
$T=1.0$, while final accuracy is highest at $T=0.3$. On name-valued questions,
moving from $T=0.3$ to $1.0$ raises coverage from $26.7\%$ to $32.5\%$ but
changes final accuracy from $21.7\%$ to $19.2\%$. The broader cohort shows
the same contrast: coverage rises from $24.1\%$ to $31.8\%$, while accuracy
changes from $17.6\%$ to $16.5\%$. Figure~\ref{fig:temperature} compares $T=0.3$
and $T=1$ on the prospective questions with new samples.

\section{Recall Readouts for Budget Allocation}
\label{app:adaptive-budget}

We test whether the readouts of Appendix~\ref{app:readout-coverage} can guide additional sampling. Extracting them takes $2.48$ and $1.73$ seconds for $489$ entities in Gemma and Llama, excluding checkpoint loading and warmup. Excluding
direction-fit overlap leaves $521$ eligible Gemma questions over $435$
entity clusters; Llama retains $598$ questions over $489$ clusters.

For each question, $256$ fixed random permutations of the retained pool define
nested candidate budgets. A policy allocates either $4$ or up to $16$
candidates, with $P(\mathrm{True})$ used for plurality tie-breaking.
We compare uniform budgets of $7$, $10$, and $13$ candidates, capped at each pool's
size (the mean budgets in Table~\ref{tab:adaptive-budget}), against random allocation,
low-readout-first allocation, and two learned allocators. The learned target
is the gain in accuracy from budget $4$ to $16$. One predictor uses
relation and entity type; the other adds the readout and its square.
Both use ridge regularization $\alpha=10$ and five entity-disjoint folds.
Encoders, standardization, and coefficients are fitted only on the other
folds. No hyperparameter or allocation rule is selected from test outcomes.

Every policy matches the uniform policy's expected candidate count within each
test fold: ranked policies split one boundary question at random, random allocation
uses a calibrated upgrade probability, and caps are applied before matching.
Paired entity-clustered intervals condition on the fitted policies and the saved candidate outcomes.

\begin{table}[t]
\centering
\small
\setlength{\tabcolsep}{3pt}
\caption{\textbf{Recall-guided allocation at matched candidate budgets.}
Percent accuracy with verification-based plurality. The final column is the
readout allocator's gain over uniform in points ($95\%$ interval).
R+T denotes relation/type; S adds the readout. Risk allocates extra
candidates to the questions with the lowest readout scores first.}
\label{tab:adaptive-budget}
\begin{tabular}{llrrrrrl}
\toprule
Model & Budget & Uniform & Random & Risk & R+T & R+T+S & Difference [95\% CI] \\
\midrule
Gemma & $7.00$ & $25.49$ & $24.92$ & $24.17$ & $25.70$ & $25.42$ & $-0.07\ [-0.86,0.74]$ \\
 & $9.79$ & $26.61$ & $26.07$ & $24.83$ & $26.89$ & $26.63$ & $+0.02\ [-0.74,0.76]$ \\
 & $11.86$ & $27.09$ & $26.92$ & $26.69$ & $26.95$ & $26.87$ & $-0.23\ [-0.72,0.21]$ \\
\midrule
Llama & $7.00$ & $44.52$ & $43.73$ & $43.69$ & $44.15$ & $44.27$ & $-0.25\ [-1.03,0.54]$ \\
 & $9.99$ & $44.88$ & $44.14$ & $43.80$ & $44.48$ & $44.73$ & $-0.15\ [-0.95,0.65]$ \\
 & $12.88$ & $45.04$ & $44.54$ & $44.06$ & $44.59$ & $45.09$ & $+0.05\ [-0.56,0.67]$ \\
\bottomrule
\end{tabular}
\end{table}

\begin{figure}[t]
\centering
\captionsetup[subfigure]{font=small,labelfont=bf,justification=raggedright,
singlelinecheck=false,skip=4pt}
\begin{subfigure}[t]{0.49\linewidth}
\centering
\includegraphics[width=\linewidth]{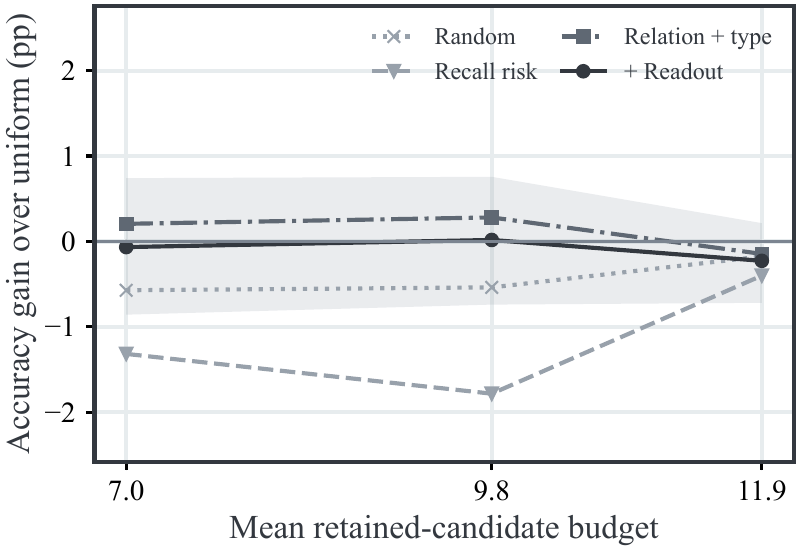}
\caption{\textbf{Gemma.} $521$ questions.}
\end{subfigure}\hfill
\begin{subfigure}[t]{0.49\linewidth}
\centering
\includegraphics[width=\linewidth]{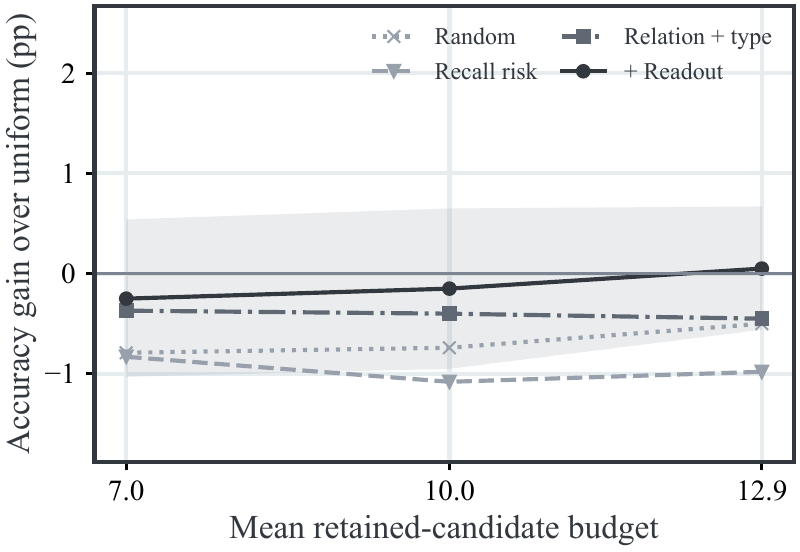}
\caption{\textbf{Llama.} $598$ questions.}
\end{subfigure}
\caption{\textbf{Recall-guided allocation at matched candidate budgets.}
Gains are relative to uniform allocation. Shading shows the paired $95\%$
interval for the readout-augmented allocator.}
\label{fig:adaptive-budget}
\end{figure}

With the readout added, accuracy stays within $0.25$ points of uniform
allocation at every tested budget, and all six intervals include zero; it differs
from the relation/type predictor by $-0.28$ to $+0.50$ points. Prioritizing
the lowest readout scores yields lower point estimates than uniform allocation
in both models (Figure~\ref{fig:adaptive-budget}). The readouts thus identify
which questions sampling will cover but not where additional candidates change
the selected answer.

\section{Measured Inference Budgets}
\label{app:measured-runtime}

Gemma and Llama each draw $32$ new samples for each of the same $600$ questions with the settings of Appendix~\ref{app:scoring}, so Gemma sampling here, unlike the primary Gemma cohort, has no top-$k$ truncation. Here $K$ counts raw
generations rather than retained candidates: each budget draws $K$ samples and
then applies the format filter, so rejected generations consume budget. Every
question stays in the denominator ($600$ questions over $489$ entity clusters
in each model), and an empty filtered pool counts as uncovered and incorrect.
We average $128$ nested subset draws at
$K\in\{1,2,4,8,16,32\}$, using the stricter criterion and plurality
with either likelihood or original-prompt verification tie-breaking.

Timing uses an RTX A5000, with float32 for Gemma and bfloat16 for Llama, on
$64$ questions chosen by a hash of question identity, independent of labels.
Each timing question receives budgets $4$, $8$, $16$, and $32$ twice, in
interleaved order and after warm-up. CUDA-synchronized wall times cover
tokenization, transfers, model calls, and score extraction, and exclude
checkpoint loading, labeling, selection, and file writes. The likelihood
pipeline includes generation and teacher-forced rescoring of the decoded text;
verification adds one evaluation per candidate. A sensitivity analysis
subtracts teacher-forcing time from both pipelines, as if likelihood were
computed during generation.

To match likelihood at $K=16$, we mix the two adjacent verification budgets
whose mean times on the timing questions bracket that target, with a mixing
probability set from timing alone, without selection labels. Accuracy averages
over all $600$ questions, and paired entity-clustered intervals condition on the
timing weights and the subset averages.

\begin{table}[H]
\centering
\small
\setlength{\tabcolsep}{5pt}
\caption{\textbf{Coverage, accuracy, and measured time.} Accuracies are
percentages; costs are seconds per question in the implemented pipeline.}
\label{tab:measured-runtime}
\begin{tabular}{lrrrrrr}
\toprule
Model & $K$ & Coverage & Likelihood accuracy & Verification & Likelihood time & Verification time \\
\midrule
Gemma & $4$ & $24.80$ & $18.90$ & $21.03$ & $0.849$ & $0.969$ \\
 & $8$ & $31.12$ & $20.66$ & $24.17$ & $0.992$ & $1.211$ \\
 & $16$ & $36.45$ & $21.97$ & $26.31$ & $1.289$ & $1.717$ \\
 & $32$ & $41.33$ & $23.67$ & $28.33$ & $1.887$ & $2.735$ \\
\midrule
Llama & $4$ & $49.80$ & $40.64$ & $42.76$ & $0.776$ & $0.865$ \\
 & $8$ & $54.42$ & $41.83$ & $43.74$ & $0.876$ & $1.029$ \\
 & $16$ & $58.95$ & $42.76$ & $44.01$ & $1.089$ & $1.399$ \\
 & $32$ & $63.17$ & $43.50$ & $43.30$ & $1.490$ & $2.099$ \\
\bottomrule
\end{tabular}
\end{table}

Without cost matching, the $K=16$ comparison replicates the primary result on
these new samples: verification raises Gemma plurality accuracy from $21.97\%$
to $26.31\%$ ($+4.34$ points, $[2.85,5.89]$) under the stricter criterion and by
$0.95$ points ($[-0.36,2.29]$) under reference matching; the Llama gain is
$1.25$ points ($[-0.38,2.91]$).

\begin{figure}[H]
\centering
\captionsetup[subfigure]{font=small,labelfont=bf,justification=raggedright,
singlelinecheck=false,skip=4pt}
\begin{subfigure}[t]{0.49\linewidth}
\centering
\includegraphics[width=\linewidth]{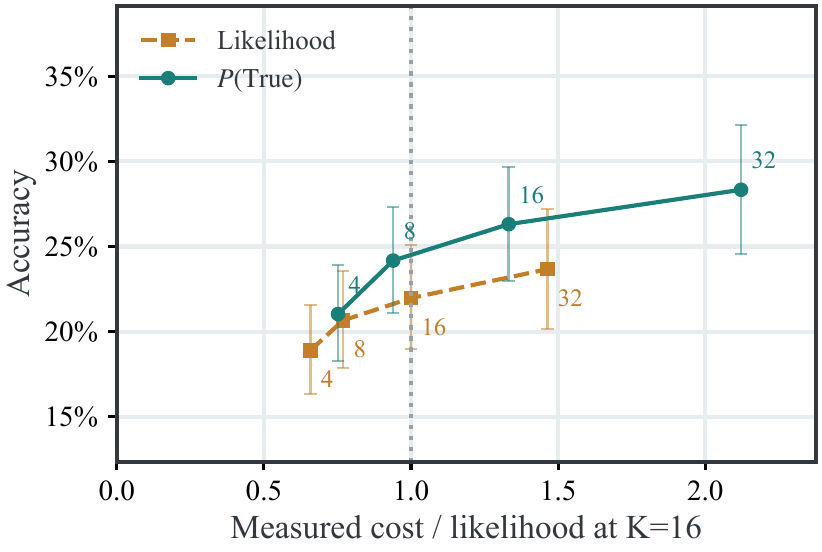}
\caption{\textbf{Gemma.} $600$ questions.}
\end{subfigure}\hfill
\begin{subfigure}[t]{0.49\linewidth}
\centering
\includegraphics[width=\linewidth]{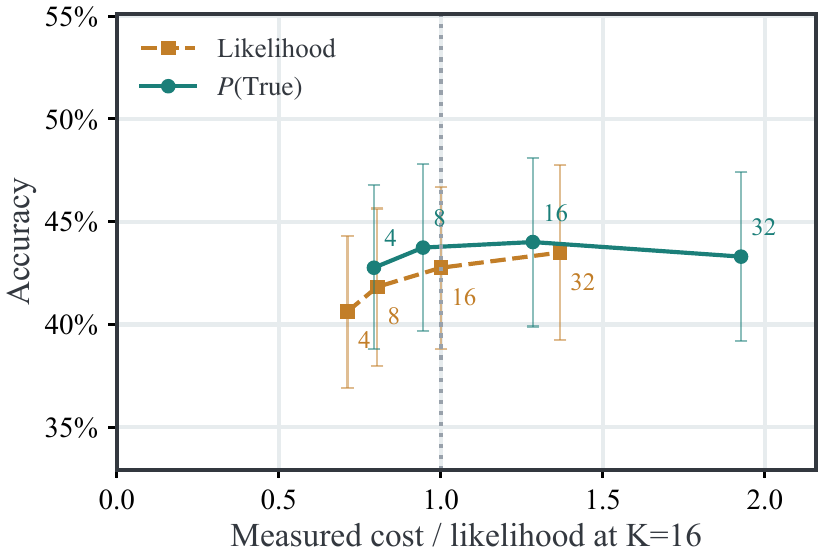}
\caption{\textbf{Llama.} $600$ questions.}
\end{subfigure}
\caption{\textbf{Accuracy at measured inference cost.} Labels mark candidate
budgets; bars show $95\%$ intervals. Cost is normalized to likelihood at $K=16$.}
\label{fig:measured-runtime}
\end{figure}

For Gemma, verification at budgets $8$ and
$16$, with probability $0.155$ on the latter, matches likelihood's $K=16$ cost
of $1.289$ seconds; accuracy is $24.50\%$ versus $21.97\%$, a $2.54$-point gain
with interval $[1.19,3.92]$, so verification improves Gemma accuracy at matched
time in the implemented pipeline. For Llama, the corresponding probability is
$0.162$ at a cost of $1.089$ seconds; accuracy is $43.78\%$ versus $42.76\%$, a
$1.02$-point gain with interval $[-0.40,2.46]$
(Figure~\ref{fig:measured-runtime}).

Crediting likelihood as computed during generation moves both matches to
mixtures of $K=4$ and $K=8$, with differences of $-0.67$ points
($[-1.90,0.55]$) in Gemma and $+0.43$ points ($[-0.80,1.70]$) in Llama, and both
intervals include zero.

\begin{figure}[H]
\centering
\begin{minipage}[c]{0.47\linewidth}
\centering
\includegraphics[width=\linewidth]{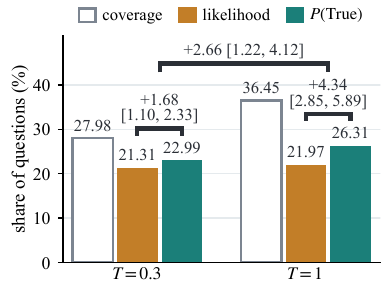}
\end{minipage}\hfill
\begin{minipage}[c]{0.49\linewidth}
\caption{\textbf{Higher temperature raises coverage and the verification gain.}
Gemma, $600$ questions with new samples, raw budget $K=16$, stricter criterion,
plurality selection. Lower brackets give the gain of $P(\mathrm{True})$ at each
temperature, and the top bracket the difference between the two gains, with
paired $95\%$ intervals.}
\label{fig:temperature}
\end{minipage}
\end{figure}

Gemma repeats the $K=16$ comparison with new samples at $T=0.3$
(Figure~\ref{fig:temperature}).

\section{Readout Prediction of Coverage}
\label{app:readout-coverage}

This analysis applies the Gemma layer-$16$ and Llama layer-$15$ readouts, with their original centering and scaling, and the fixed Qwen-Scope
feature of Section~\ref{sec:setup} to the selection questions. The Gemma and
Llama scores are read at the final entity token of the entity-only prompt and
assigned to every question about that entity. The Qwen3 feature is read at the
final entity mention of each question prompt, so it can differ between questions
about the same entity but not between a question's candidates. No readout is
refitted on the selection questions. Gemma excludes the $20$ questions whose
entities were used to fit its direction, leaving $521$ questions over $435$
entities and $6{,}595$ candidates. Llama retains all $598$ eligible questions
over $489$ entities and $9{,}349$ candidates. Qwen3's candidate pools for the
same questions use the sampling settings of the Llama study ($16$ samples per question; Appendix~\ref{app:scoring}). After filtering, $501$ questions
keep all $16$ samples and the other $99$ keep $8$--$15$, so all $600$ questions
over $489$ entities are eligible, with $9{,}443$ candidates. The target is
whether the full retained pool of up to $16$ candidates contains an answer
accepted by the stricter criterion; $192$ Gemma, $355$ Llama, and $244$ Qwen3
questions do.

AUROC compares covered with uncovered questions and gives half credit to ties.
Across all questions it is $0.7815$ ($95\%$ CI $[0.7385,0.8214]$) in Gemma,
$0.8165$ ($[0.7790,0.8526]$) in Llama, and $0.6898$ ($[0.6457,0.7313]$) in Qwen3. For Gemma
and Llama, the within-relation AUROC compares only
questions of the same relation and weights each relation by its number of
covered--uncovered pairs; it is $0.8512$ in Gemma and $0.8715$ in Llama, and
the unweighted relation average is $0.7898$ and $0.8588$; in Qwen3, the
within-relation AUROC is $0.6632$. Intervals use $3{,}000$
entity-clustered bootstrap draws with seed $20260924$. Quintiles split the
entities into groups of $87$ in Gemma and $97$--$98$ in Llama, and each includes
all eligible questions about its entities (Table~\ref{tab:readout-coverage}). In Gemma, the two highest quintiles have similar coverage. Because the Qwen3
readout can differ between relations of the same entity, its quintiles split the
$600$ questions into groups of $120$; coverage is $27.5\%$ ($33/120$) in the lowest
and $72.5\%$ ($87/120$) in the highest.

\begin{table}[H]
\centering
\small
\setlength{\tabcolsep}{5pt}
\caption{\textbf{Coverage by readout quintile.} Stricter criterion, full
retained pools. Coverage is in percent, with $95\%$ entity-clustered
intervals.}
\label{tab:readout-coverage}
\begin{tabular}{llrrrl}
\toprule
Model & Quintile & Entities & Questions & Covered & Coverage [95\% CI] \\
\midrule
Gemma & 1 (lowest) & $87$ & $99$ & $11$ & $11.1$ $[5.3,17.8]$ \\
 & 2 & $87$ & $104$ & $17$ & $16.3$ $[9.1,24.2]$ \\
 & 3 & $87$ & $112$ & $30$ & $26.8$ $[18.6,35.9]$ \\
 & 4 & $87$ & $102$ & $68$ & $66.7$ $[57.6,75.2]$ \\
 & 5 (highest) & $87$ & $104$ & $66$ & $63.5$ $[54.2,72.7]$ \\
\midrule
Llama & 1 (lowest) & $98$ & $121$ & $19$ & $15.7$ $[9.8,22.3]$ \\
 & 2 & $98$ & $117$ & $47$ & $40.2$ $[31.1,49.5]$ \\
 & 3 & $98$ & $119$ & $88$ & $73.9$ $[64.8,82.8]$ \\
 & 4 & $98$ & $117$ & $92$ & $78.6$ $[71.5,85.8]$ \\
 & 5 (highest) & $97$ & $124$ & $109$ & $87.9$ $[82.1,93.4]$ \\
\bottomrule
\end{tabular}
\end{table}

Among covered questions, the
readout predicts whether plurality returns a correct answer with AUROC $0.650$
($[0.565,0.730]$) under the likelihood tie-break and $0.662$
($[0.570,0.746]$) under the $P(\mathrm{True})$ tie-break in Gemma, and $0.619$
($[0.549,0.687]$) and $0.622$ ($[0.555,0.689]$) in Llama; tie-averaged
selections enter as fractional labels.

\paragraph{Nested comparisons.} To separate the readout from question
difficulty, we compare $M_0$, which contains the exposure, name, type, and
relation features of Section~\ref{sec:recall-eval} and a question-level
common-answer prior, with $M_1=M_0+S$. Qwen3's $M_0$ omits name likelihood; its
only name feature is the length in characters. The prior is the frequency of the
question's reference answer among other entities' reference answers to the same
relation, computed leave-one-entity-out; a relation-level prior would be
redundant with the relation indicators. Because it uses reference answers, the
prior controls for benchmark-side difficulty and is not available at inference.
The first target is $K=16$ coverage on all eligible questions; the second is
plurality success on covered questions, under each tie-break score. $M_0$ and $M_1$ are both L2-regularized logistic regressions, cross-fitted over five entity-disjoint
folds: preprocessing and regularization are fitted on the training folds, and
each question receives an out-of-fold prediction. Log loss is the primary measure, with Brier score and AUROC as secondary measures; intervals use
$10{,}000$ entity-clustered bootstrap draws (Table~\ref{tab:readout-nested}). The low-prior check computes the raw
readout AUROC for coverage within the six low-prior relations, whose common
answers cover $2.9\%$ of questions: $301$ Gemma and $315$ Llama questions, on
which the readout's AUROC is $0.807$ and $0.861$. For Qwen3,
it reports the nested improvement within the same relations ($315$ questions).

\begin{table}[H]
\centering
\small
\setlength{\tabcolsep}{5pt}
\caption{\textbf{Readout information beyond question difficulty.} Held-out
improvement from adding the readout to $M_0$ (exposure, name, type, and
relation features and a common-answer prior; no name likelihood for Qwen3), from
out-of-fold predictions over five entity-disjoint folds, with $95\%$
entity-clustered intervals. Coverage is at $K=16$; success is plurality success
on covered questions under the named tie-break. The Qwen3
low-prior row uses the six low-prior relations. AUROC runs from $M_0$ to $M_1$.}
\label{tab:readout-nested}
\setlength{\tabcolsep}{1.9pt}
\begin{tabular}{llrlll}
\toprule
Model & Target & $n$ & $\Delta L$ [95\% CI] & $\Delta$Brier [95\% CI] & AUROC \\
\midrule
Gemma & Coverage & $521$ & $+0.013$ $[0.005,0.021]$ & $+0.005$ $[0.002,0.008]$ & $0.750\rightarrow0.790$ \\
 & Success, likelihood & $192$ & $+0.0061$ $[-0.0004,0.0128]$ & $+0.0022$ $[-0.0002,0.0046]$ & $0.658\rightarrow0.673$ \\
 & Success, $P(\mathrm{True})$ & $192$ & $+0.0030$ $[-0.0028,0.0089]$ & $+0.0011$ $[-0.0010,0.0033]$ & $0.667\rightarrow0.675$ \\
\midrule
Llama & Coverage & $598$ & $+0.017$ $[0.007,0.027]$ & $+0.006$ $[0.002,0.010]$ & $0.792\rightarrow0.833$ \\
 & Success, likelihood & $355$ & $+0.0037$ $[-0.0008,0.0082]$ & $+0.0014$ $[-0.0003,0.0031]$ & $0.625\rightarrow0.636$ \\
 & Success, $P(\mathrm{True})$ & $355$ & $+0.0020$ $[-0.0020,0.0061]$ & $+0.0007$ $[-0.0007,0.0023]$ & $0.629\rightarrow0.635$ \\
\midrule
Qwen3 & Coverage & $600$ & $+0.0137$ $[-0.0012,0.0276]$ & $+0.0032$ $[-0.0022,0.0085]$ & $0.862\rightarrow0.870$ \\
 & Low prior & $315$ & $+0.0246$ $[0.0050,0.0468]$ & $+0.0054$ $[-0.0018,0.0127]$ & $0.793\rightarrow0.814$ \\
  & Success, likelihood & $244$ & $+0.0039$ $[-0.0015,0.0093]$ & $+0.0014$ $[-0.0005,0.0033]$ & $0.726\rightarrow0.739$ \\
  & Success, $P(\mathrm{True})$ & $244$ & $-0.0001$ $[-0.0047,0.0045]$ & $0.0000$ $[-0.0016,0.0016]$ & $0.731\rightarrow0.732$ \\
\bottomrule
\end{tabular}
\end{table}

In Qwen3, adding the readout raises plurality-success AUROC by $0.0133$
($[-0.004,0.031]$) under the likelihood tie-break and by $0.0009$
($[-0.015,0.017]$) under the $P(\mathrm{True})$ tie-break.

On the same questions, verification ranks candidates better than likelihood.
In Gemma, pair-weighted AUROC rises from $0.678$ to $0.797$ over $189$ mixed
questions, and plurality accuracy rises from $22.3\%$ to $27.3\%$; these values
differ slightly from the primary $541$-question values in Table~\ref{tab:ptrue}
because $20$ questions are excluded here. The Llama and Qwen3 cohorts are their full selection cohorts, so their values are
those in Appendices~\ref{app:llama-selection} and~\ref{app:qwen-selection}.

\section{Ranking and Selection in Additional Models}
\label{app:qwen-selection}

\subsection{Qwen3-8B}

The Qwen3-8B selection study uses the checkpoint of the Qwen3 recall study
(Appendix~\ref{app:qwen}) and the same $600$ questions as the prospective Gemma
cohort. Responses are sampled from bare-format prompts, $16$ per question, with the settings of Appendix~\ref{app:scoring}, and the format filter and both correctness criteria are applied unchanged. All $600$ questions
over $489$ entity clusters retain at least eight candidates, and $501$ retain all
$16$, giving $9{,}443$ candidates. Likelihood is computed by teacher forcing,
and verification uses the original Yes/No prompt with Qwen3's thinking mode
disabled.

\begin{table}[H]
\centering
\small
\setlength{\tabcolsep}{5pt}
\caption{\textbf{Qwen3-8B ranking and selection.} Stricter criterion, $600$
questions, $K=16$; $244$ questions ($40.67\%$) are covered. AUROC uses the
$168$ mixed questions ($5{,}707$ correct--incorrect pairs). Accuracies and
conversion are percentages; accuracy differences are points.}
\label{tab:qwen-selection}
\begin{tabular}{lrrl}
\toprule
Measure & Likelihood & $P(\mathrm{True})$ & Difference [95\% CI] \\
\midrule
Pair-weighted AUROC & $0.6206$ & $0.6984$ & $+0.0778$ $[0.038,0.118]$ \\
Query-macro AUROC & $0.6269$ & $0.6749$ & $+0.0480$ $[0.0040,0.0921]$ \\
Argmax selection & $26.17$ & $27.33$ & $+1.17$ $[-0.91,3.27]$ \\
Plurality selection & $26.17$ & $27.33$ & $+1.17$ $[-0.75,3.10]$ \\
Conversion (plurality) & $64.34$ & $67.21$ & --- \\
\bottomrule
\end{tabular}
\end{table}

Verification ranks answers to the same question better than likelihood:
pair-weighted AUROC rises by $0.078$ ($[0.038,0.118]$) and query-macro AUROC by
$0.048$ ($[0.0040,0.0921]$). Under both selection rules, verification
returns $164$ correct answers against $157$ ($+1.17$ points), and both intervals
include zero.

\subsection{Gemma 2 9B and 27B}
\label{app:gemma27b}

Gemma~2 9B-IT and 27B-IT \citep{gemmateam2024gemma2} run in bfloat16 on the same
$600$ questions, of which $581$ and $564$ are eligible. Every run, including the float32 run of 9B below, samples $16$ responses per question from the bare prompt with the settings of the prospective Gemma cohort: $T=1$, top-$k=50$, no nucleus truncation, and a $24$-token cap (Appendix~\ref{app:scoring}). Tables~\ref{tab:gemma9b-selection}
and~\ref{tab:gemma27b-selection} give their ranking and selection results. Under the stricter criterion, $309$ of the $581$
9B questions ($53.18\%$) and $332$ of the $564$ 27B questions ($58.87\%$) are
covered, and likelihood already converts $81.88\%$ and $81.63\%$ of them, against
$61.27\%$ in Gemma~2 2B-IT (Table~\ref{tab:ptrue}), and the selection gain is also smaller. A float32 run of 9B, with $577$
eligible questions, gives nearly the same coverage: $307$ ($53.21\%$) under the
stricter criterion and $313$ ($54.25\%$) under reference matching, against $309$
($53.18\%$) and $321$ ($55.25\%$) in bfloat16. In bfloat16, some $P(\mathrm{True})$ scores tie exactly, and ties
are averaged uniformly (Section~\ref{sec:setup}). In 27B, one question has a
correct and an incorrect candidate tied for plurality selection, which gives
$274.5$ correct answers ($48.67\%$), and $57$ of the $3{,}978$ correct--incorrect
pairs ($1.43\%$) are tied; resolving every tie in either direction gives a
plurality gain of $0.53$ or $0.71$ points and changes pair-weighted AUROC by at
most $0.0072$.

\begin{table}[H]
\centering
\small
\setlength{\tabcolsep}{5pt}
\caption{\textbf{Gemma~2 9B-IT ranking and selection.} Stricter criterion, $581$
eligible questions, $K=16$; $309$ questions ($53.18\%$) are covered. Accuracies and
conversion are percentages; accuracy differences are points.}
\label{tab:gemma9b-selection}
\begin{tabular}{lrrl}
\toprule
Measure & Likelihood & $P(\mathrm{True})$ & Difference [95\% CI] \\
\midrule
Pair-weighted AUROC & $0.700$ & $0.732$ & $+0.032$ $[0.006,0.058]$ \\
Argmax selection & $43.55$ & $43.89$ & $+0.34$ $[-0.91,1.59]$ \\
Plurality selection & $43.55$ & $43.72$ & $+0.17$ $[-0.86,1.20]$ \\
Conversion (plurality) & $81.88$ & $82.20$ & --- \\
Option-list selections & $0$ & $0$ & --- \\
\bottomrule
\end{tabular}
\end{table}

\begin{table}[H]
\centering
\small
\setlength{\tabcolsep}{5pt}
\caption{\textbf{Gemma~2 27B-IT ranking and selection.} Stricter criterion, $564$
eligible questions, $K=16$; $332$ questions ($58.87\%$) are covered. Accuracies
and conversion are percentages; the accuracy difference is in points. Argmax
selection and option-list counts, reported for Gemma~2 9B-IT in
Table~\ref{tab:gemma9b-selection}, are not available for this run.}
\label{tab:gemma27b-selection}
\begin{tabular}{lrrl}
\toprule
Measure & Likelihood & $P(\mathrm{True})$ & Difference [95\% CI] \\
\midrule
Pair-weighted AUROC & $0.643$ & $0.673$ & $+0.030$ $[-0.033,0.091]$ \\
Plurality selection & $48.05$ & $48.67$ & $+0.62$ $[-1.03,2.28]$ \\
Conversion (plurality) & $81.63$ & $82.68$ & --- \\
\bottomrule
\end{tabular}
\end{table}

\subsection{Qwen3.5, Qwen3.6, and Qwen3.8}
\label{app:qwen-family}

Four further Qwen models \citep{qwen35,qwen36,qwen38}
sample the same $600$ questions at $T=1$ without top-$k$
truncation, $16$ raw samples per question, and are scored under the stricter
criterion (Table~\ref{tab:qwen-family}). The checkpoints are
\path{Qwen/Qwen3.5-2B}, \path{Qwen/Qwen3.5-4B}, \path{Qwen/Qwen3.6-27B}, and
\path{Qwen/Qwen3.8-27B}, and verification runs with thinking disabled. Verification raises plurality conversion
in every model except Qwen3.5-2B, where it falls from $58.65\%$ to $50.48\%$ and
pair-weighted AUROC changes by only $0.008$; Table~\ref{tab:models} gives the
ranking and plurality differences with their intervals,
Table~\ref{tab:qwen-family-ranking} adds query-macro AUROC and argmax selection,
and Appendix~\ref{app:rescoring} compares the two 27B models' scores with the
generative baselines and entity masking. In Qwen3.5-4B, the plurality interval
reaches zero, while the argmax and both AUROC intervals lie above zero.

\begin{table}[H]
\centering
\small
\setlength{\tabcolsep}{4pt}
\caption{\textbf{Coverage and conversion across Qwen models.} Stricter criterion,
$T=1$, no top-$k$ truncation, $16$ raw samples per question. Conversion is the
share of covered questions that selection answers correctly, with likelihood or
$P(\mathrm{True})$; retention is the share of the $9{,}600$ samples kept by the
format filter. All values except eligible counts are percentages.}
\label{tab:qwen-family}
\begin{tabular}{lrrrrrr}
\toprule
 & & & \multicolumn{3}{c}{Conversion} & \\
\cmidrule(lr){4-6}
Model & Eligible & Coverage & \shortstack{Likelihood\\argmax} & \shortstack{Likelihood\\plurality} & \shortstack{$P(\mathrm{True})$\\plurality} & Retention \\
\midrule
Qwen3.5-2B & $528$ & $19.70$ & $59.62$ & $58.65$ & $50.48$ & $65.85$ \\
Qwen3.5-4B & $565$ & $33.45$ & $57.67$ & $58.20$ & $64.55$ & $73.49$ \\
Qwen3-8B & $600$ & $40.67$ & $64.34$ & $64.34$ & $67.21$ & $98.36$ \\
Qwen3.6-27B & $599$ & $48.58$ & $62.54$ & $61.86$ & $68.73$ & $89.51$ \\
Qwen3.8-27B & $594$ & $41.25$ & $60.41$ & $60.00$ & $68.23$ & $81.49$ \\
\bottomrule
\end{tabular}
\end{table}

\begin{table}[H]
\centering
\small
\setlength{\tabcolsep}{4pt}
\caption{\textbf{Ranking and argmax selection in the newer Qwen models.} Stricter
criterion. Arrows run from mean log-likelihood to $P(\mathrm{True})$; differences
are $P(\mathrm{True})$ minus mean log-likelihood, with paired $95\%$
entity-clustered intervals below. Pair-weighted and query-macro differences have
separate intervals; plurality results are in Table~\ref{tab:models}.}
\label{tab:qwen-family-ranking}
\begin{tabular}{lcrrcr}
\toprule
Model & \shortstack{Pair-weighted\\AUROC} & $\Delta$ & \shortstack{Query-macro\\$\Delta$} & \shortstack{Argmax\\accuracy (\%)} & \shortstack{Gain\\(points)} \\
\midrule
Qwen3.5-2B & $0.605{\to}0.613$ & $+0.008$ & $+0.012$ & $11.74{\to}9.94$ & $-1.80$ \\
 & & $[-0.053,0.069]$ & $[-0.049,0.073]$ & & $[-3.53,-0.10]$ \\[2pt]
Qwen3.5-4B & $0.617{\to}0.755$ & $+0.139$ & $+0.131$ & $19.29{\to}21.59$ & $+2.30$ \\
 & & $[0.101,0.176]$ & $[0.090,0.171]$ & & $[0.18,4.48]$ \\[2pt]
Qwen3.6-27B & $0.612{\to}0.792$ & $+0.180$ & $+0.198$ & $30.38{\to}33.72$ & $+3.34$ \\
 & & $[0.149,0.213]$ & $[0.162,0.235]$ & & $[0.87,5.87]$ \\[2pt]
Qwen3.8-27B & $0.625{\to}0.773$ & $+0.149$ & $+0.159$ & $24.92{\to}28.48$ & $+3.56$ \\
 & & $[0.111,0.186]$ & $[0.118,0.199]$ & & $[1.13,6.06]$ \\
\bottomrule
\end{tabular}
\end{table}

\section{Semantic Judge and Human Labels}
\label{app:semantic}

\paragraph{Protocol.}
The semantic labeling protocol was specified before the judge was run and before any
selector was compared under its labels. The judge is GPT-5.6 Sol
\citep{openai2026gpt56}, API model \texttt{gpt-5.6-sol}, queried in September 2026
with reasoning effort set to none, temperature $0$, and top-$p$ $1$. It runs in two
passes, and neither pass sees model scores, selector identities, or labels from reference
matching or the stricter criterion. Pass~0 checks each question's references once, from the question
and its reference answers alone, without any candidate, and returns
\texttt{VALID} or \texttt{INVALID\_OR\_AMBIGUOUS}; an author checks every flagged
question, and confirmed invalid questions leave the semantic denominator. Pass~1
labels each distinct combination of question, reference set, and exact candidate
text once, reuses the label for every occurrence, and returns \texttt{CORRECT} or
\texttt{INCORRECT}. Aliases, paraphrases, and enumerations that end by choosing
the correct answer are accepted; unresolved alternatives, contradictions,
refusals, and mixtures of correct and incorrect assertions are rejected; hedging
alone does not make an answer incorrect. For multi-valued relations, one valid
value suffices unless the question asks for an exhaustive set, and every value the
response asserts must be correct: for the reference set \emph{The Beatles; Wings},
``The Beatles'' and ``The Beatles and Wings'' are correct, whereas ``The Beatles and
Queen'' and ``either The Beatles or Queen'' are not. The prompt was developed on the
earlier 400-candidate audit (Appendix~\ref{app:human-audit}) and fixed before the full relabeling. Appendix~\ref{app:prompts} reproduces both prompts.
\paragraph{Validation.}
Validation uses a new blinded set of outputs labeled after the
protocol was fixed: every output that argmax or plurality selects differently
under the two scores, including tied candidates with nonzero selection weight
($500$ outputs from $220$ argmax and $80$ plurality disagreement questions), plus $250$ random candidates and $200$ candidates from hard strata (option lists, negations,
aliases, hedged answers, refusals). Three authors label each output as correct or incorrect without seeing automatic labels, model scores, or selector
identities. The human label is the majority vote of the three blinded annotators;
with three binary labels, every item has a unique majority label. On the $250$ random candidates, the judge agrees with the human labels on
$233$ ($93.2\%$; Cohen's $\kappa=0.84$); on the outputs selected by likelihood and by verification, agreement is $92.5\%$
and $93.7\%$ (difference $-1.2$ points, $[-3.6,1.3]$).

\begin{table}[H]
\centering
\small
\setlength{\tabcolsep}{4pt}
\caption{\textbf{Selection gains under four labelings.} Gemma, $K=16$, all $541$
questions (every reference passes the validity check). Pair-weighted AUROC is shown from likelihood
to $P(\mathrm{True})$; gains are points with paired $95\%$ entity-clustered
intervals. Human labels cover every output on which the selectors differ, so the
human gains apply to the whole cohort.}
\label{tab:semantic-endpoints}
\begin{tabular}{lllll}
\toprule
Labels & Coverage & Pair-weighted AUROC & Argmax gain & Plurality gain \\
\midrule
Reference & $38.82\%$ & $0.711\rightarrow0.733$ & $+0.74$ $[-1.71,3.19]$ & $+1.29$ $[-0.8,3.4]$ \\
Stricter & $37.71\%$ & $0.680\rightarrow0.782$ & $+5.36$ $[2.56,8.19]$ & $+4.99$ $[2.7,7.3]$ \\
Semantic & $38.63\%$ & $0.701\rightarrow0.778$ & $+4.07$ $[1.6,6.5]$ & $+3.88$ $[1.6,6.1]$ \\
Human & --- & --- & $+4.25$ $[1.63,6.87]$ & $+3.70$ $[1.31,6.08]$ \\
\midrule
Semantic $-$ reference & --- & --- & $+3.33$ $[1.2,5.4]$ & $+2.59$ $[0.9,4.3]$ \\
Stricter $-$ semantic & --- & --- & $+1.29$ $[0.28,2.36]$ & $+1.11$ $[0.21,2.05]$ \\
\bottomrule
\end{tabular}
\end{table}

\begin{table}[H]
\centering
\small
\setlength{\tabcolsep}{4pt}
\caption{\textbf{Reference-matching errors by selector.} Errors are selected outputs whose label under reference matching differs from the semantic label, as percentages of the $541$
questions, with paired $95\%$ entity-clustered intervals. The last column counts the
errors that are option lists.}
\label{tab:reference-errors}
\begin{tabular}{lllll}
\toprule
Selection & Likelihood & $P(\mathrm{True})$ & Difference & Option lists \\
\midrule
Argmax & $3.51$ $[1.95,5.15]$ & $0.55$ $[0.05,1.30]$ & $+2.96$ $[1.40,4.55]$ & $18/19$ vs.\ $1/3$ \\
Plurality & $2.22$ $[1.15,3.84]$ & $0.37$ $[0.00,0.95]$ & $+1.85$ $[0.58,3.16]$ & $12/12$ vs.\ $0/2$ \\
\bottomrule
\end{tabular}
\end{table}

\paragraph{Readout analysis under semantic labels.}
Table~\ref{tab:semantic-readout} repeats the nested comparisons of
Appendix~\ref{app:readout-coverage} with semantic coverage and success labels.
The relation groups and the common-answer prior are built from other entities'
reference answers, so they do not depend on candidate labels and are unchanged.
Semantic coverage rises from $12.1\%$ ($12/99$) in the lowest readout quintile to
$64.4\%$ ($67/104$) in the highest in Gemma, and from $15.7\%$ ($19/121$) to
$87.1\%$ ($108/124$) in Llama; in the six low-prior relations, the
raw coverage AUROC is $0.800$ ($[0.74,0.86]$) and $0.853$ ($[0.80,0.90]$).

\begin{table}[H]
\centering
\small
\setlength{\tabcolsep}{4pt}
\caption{\textbf{Readout information under semantic labels.} Held-out improvement
from adding the readout to $M_0$, with $95\%$ entity-clustered intervals;
AUROC is the readout's raw AUROC for coverage.}
\label{tab:semantic-readout}
\begin{tabular}{llrlll}
\toprule
Model & Target & $n$ & AUROC & $\Delta L$ [95\% CI] & $\Delta$Brier [95\% CI] \\
\midrule
Gemma & Coverage & $521$ & $0.779$ & $+0.0120$ $[0.0040,0.0202]$ & $+0.0044$ $[0.0013,0.0075]$ \\
 & Success, likelihood & $197$ & --- & $+0.0055$ $[-0.0010,0.0121]$ & $+0.0020$ $[-0.0003,0.0043]$ \\
 & Success, $P(\mathrm{True})$ & $197$ & --- & $+0.0028$ $[-0.0031,0.0087]$ & $+0.0010$ $[-0.0010,0.0031]$ \\
\midrule
Llama & Coverage & $598$ & $0.812$ & $+0.0155$ $[0.0060,0.0252]$ & $+0.0056$ $[0.0020,0.0093]$ \\
 & Success, likelihood & $359$ & --- & $+0.0033$ $[-0.0011,0.0079]$ & $+0.0012$ $[-0.0004,0.0029]$ \\
 & Success, $P(\mathrm{True})$ & $359$ & --- & $+0.0018$ $[-0.0022,0.0060]$ & $+0.0007$ $[-0.0008,0.0022]$ \\
\midrule
Qwen3 & Coverage & $600$ & $0.688$ & $+0.0128$ $[-0.0015,0.0264]$ & $+0.0046$ $[-0.0005,0.0096]$ \\
  & Low prior & $315$ & --- & $+0.0235$ $[0.0040,0.0440]$ & $+0.0085$ $[0.0014,0.0157]$ \\
  & Success, likelihood & $250$ & --- & $+0.0028$ $[-0.0015,0.0072]$ & $+0.0010$ $[-0.0005,0.0026]$ \\
  & Success, $P(\mathrm{True})$ & $250$ & --- & $+0.0014$ $[-0.0026,0.0054]$ & $+0.0005$ $[-0.0009,0.0019]$ \\
\bottomrule
\end{tabular}
\end{table}

\paragraph{Semantic selection results.} All $541$ references pass the validity
check. Under semantic labels, $209$ questions are covered and $207$ are mixed; over
the mixed questions, pair-weighted AUROC rises from $0.701$ to $0.778$ ($+0.077$,
$[0.037,0.118]$).
Likelihood and verification select correct answers on $127$ and $149$ questions
under argmax and on $134$ and $155$ under plurality, so plurality conversion rises from $64.11\%$ to $74.16\%$ (Table~\ref{tab:semantic-endpoints}).
By common-answer prior, the semantic plurality gain is $7.86$ points ($[3.55,12.10]$) in the $229$ high-prior questions ($69$ to $87$ correct, $30.13\%$ to $37.99\%$) and
$0.96$ points ($[-1.35,3.30]$) in the other $312$ ($65$ to $68$, $20.83\%$ to
$21.79\%$), a difference of $6.90$ points ($[2.00,11.75]$), against $10.04$ and $1.28$
under the stricter criterion (Table~\ref{tab:relation-groups}). Relative to the
stricter criterion, semantic labels credit nine more likelihood plurality selections
($125$ to $134$; six in high-prior relations) and three more verification
selections ($152$ to $155$), so the stricter criterion gives a plurality gain
$1.11$ points larger ($[0.21,2.05]$) and an argmax gain $1.29$ points larger
($[0.28,2.36]$). This matches the two-author audit, in which about a third of
likelihood's disputed selections are option lists that end by choosing the correct
answer (Appendix~\ref{app:human-audit}).

\paragraph{Comparison with reference matching and human labels.} Under argmax,
the gain over mean log-likelihood is $4.07$ points under semantic labels and $4.25$
under human labels, in line with the two-author audit ($3.88$ and $4.25$).
Relative to semantic labels, reference matching lowers the plurality gain by
$2.59$ points ($[0.9,4.3]$) and the argmax gain by $3.33$ ($[1.2,5.4]$), whereas
semantic and human labels differ by one question on net under both rules ($0.18$
points). Under plurality, reference matching wrongly accepts $12$ of likelihood's selections ($2.22\%$), all of
them option lists, and rejects $2$ of verification's correct selections
($0.37\%$), a difference of $1.85$ points ($[0.58,3.16]$); together, these $14$
questions make up the $2.59$-point gap between the semantic and reference gains
(Table~\ref{tab:reference-errors}).

For Llama, semantic labels give $60.0\%$ coverage and $305$ mixed questions.
Pair-weighted AUROC rises from $0.646$ to $0.759$ ($+0.113$, $[0.083,0.143]$),
argmax selection from $44.98\%$ to $46.32\%$ ($+1.34$ points), and plurality from
$44.82\%$ to $46.32\%$ ($+1.51$, $[-0.10,3.15]$).

\section{Generative Rescoring and Entity Masking}
\label{app:rescoring}

This appendix defines the scores of Section~\ref{sec:rescoring}. Let $y$ be a
candidate with surrounding whitespace removed, $q$ its question, and $\tilde q$
the same question with the entity replaced by \emph{this \{type\}}, where
\emph{\{type\}} names the entity type. Summed log-likelihood is
$\sum_t \log p(y_t\mid q,y_{<t})$ on the bare-format prompt of mean log-likelihood.
Chat-template likelihood is
\[
s_{\rm chat}(q,y)=\sum_t \log p\!\left(y_t\mid C(q),y_{<t}\right),
\]
where $C(q)$ renders the user message \emph{Return only the requested answer, without alternatives or explanation. Question: \{question\}} with the model's chat template and $y$ is the assistant
reply (Appendix~\ref{app:prompts}). The relation-prior PMI is
$s_{\rm chat}(q,y)-s_{\rm chat}(\tilde q,y)$. Masked $P(\mathrm{True})$ uses the
verification prompt with $\tilde q$ in place of $q$. Each score serves as the
plurality tie-break of Section~\ref{sec:setup}.

In Gemma under semantic labels (Figure~\ref{fig:decomposition}), plurality
selection returns a correct answer on $134$ of the $541$ questions under mean
log-likelihood, $138$ under summed log-likelihood, $144$ under chat-template
likelihood, $113$ under PMI, and $155$ under verification. On the $229$ high-prior
questions, the counts are $69$, $80$, and $87$ for mean log-likelihood, chat-template likelihood, and verification, and $81$ for
masked $P(\mathrm{True})$;
on the other $312$, they are $65$, $64$, and $68$.

\begin{figure}[H]
\centering
\captionsetup[subfigure]{font=small,labelfont=bf,justification=raggedright,
singlelinecheck=false,skip=4pt}
\begin{subfigure}[t]{0.345\linewidth}
\centering
\includegraphics[width=\linewidth]{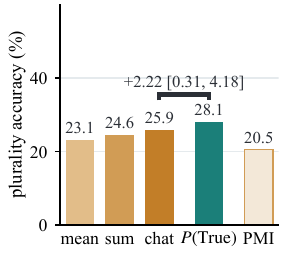}
\caption{\textbf{All questions.}}
\end{subfigure}\hfill
\begin{subfigure}[t]{0.34\linewidth}
\centering
\includegraphics[width=\linewidth]{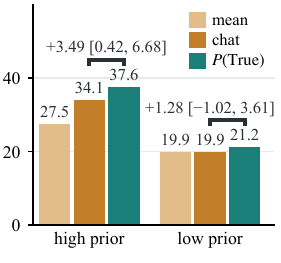}
\caption{\textbf{By common-answer prior.}}
\end{subfigure}\hfill
\begin{subfigure}[t]{0.28\linewidth}
\centering
\includegraphics[width=\linewidth]{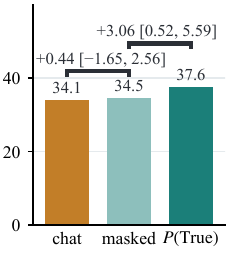}
\caption{\textbf{Entity masking, high prior.}}
\end{subfigure}
\caption{\textbf{Figure~\ref{fig:decomposition} under the stricter criterion.}
Same questions, scores, and layout. Brackets give paired differences in points
with $95\%$ intervals.}
\label{fig:decomposition-strict}
\end{figure}

Under the stricter criterion (Figure~\ref{fig:decomposition-strict}), the
corresponding counts are $125$, $133$, $140$, $111$, and $152$; $63$, $78$, $86$,
and $79$; and $62$, $62$, and $66$. Verification then exceeds chat-template
likelihood by $2.22$ points ($[0.31,4.18]$) over all questions, by $3.49$
($[0.42,6.68]$) in the high-prior questions, and by $1.28$ ($[-1.02,3.61]$) in the
others; in the high-prior questions, masked $P(\mathrm{True})$ is $0.44$ points above chat-template likelihood
($[-1.65,2.56]$) and $3.06$ points below $P(\mathrm{True})$ ($[0.52,5.59]$).

In Qwen3-8B, under the stricter criterion, neither chat-template likelihood nor PMI
improves plurality accuracy over mean log-likelihood (Table~\ref{tab:qwen-rescoring}),
and verification's plurality gain over mean log-likelihood is $1.17$ points
($[-0.75,3.10]$; Appendix~\ref{app:qwen-selection}). Six of its seven net additional
correct selections fall in the high-prior relations. There, however, chat-template
likelihood matches mean log-likelihood, whereas in Gemma it closes most of the
high-prior gap; verification's gain over chat-template likelihood is therefore
similar in the two groups, $2.11$ and $2.54$ points over all $285$ and $315$
questions.

\begin{table}[H]
\centering
\small
\setlength{\tabcolsep}{5pt}
\caption{\textbf{Qwen3-8B plurality selection by relation group.} Stricter
criterion. Correct selections, with conversion among covered questions ($165$
high-prior, $79$ low-prior, $244$ in total).}
\label{tab:qwen-rescoring}
\begin{tabular}{lrrr}
\toprule
Score & High-prior & Low-prior & All \\
\midrule
Mean log-likelihood & $109$ ($66.06\%$) & $48$ ($60.76\%$) & $157$ ($64.34\%$) \\
Chat-template likelihood & $109$ ($66.06\%$) & $41$ ($51.90\%$) & $150$ ($61.48\%$) \\
PMI & $104$ ($63.03\%$) & $37$ ($46.84\%$) & $141$ ($57.79\%$) \\
$P(\mathrm{True})$ & $115$ ($69.70\%$) & $49$ ($62.03\%$) & $164$ ($67.21\%$) \\
\bottomrule
\end{tabular}
\end{table}

In two 27B Qwen models, Qwen3.6-27B and Qwen3.8-27B, the same scores are computed on
the same candidates (Tables~\ref{tab:qwen27b-rescoring}
and~\ref{tab:qwen27b-differences}). Verification ranks candidates far better than
every generative score. Chat-template likelihood recovers $7$ and $3$ of
verification's about $20$ net additional correct selections over mean
log-likelihood, against about half in Gemma, and summed log-likelihood none; PMI
raises pair-weighted AUROC above chat-template likelihood without raising plurality
accuracy further. Masked $P(\mathrm{True})$ matches chat-template
likelihood in plurality accuracy and in pair-weighted and query-macro AUROC, and
unmasked $P(\mathrm{True})$ rescored in the same run exceeds it on all three
measures. In the
high-prior relations, masking leaves conversion almost unchanged in Qwen3.6-27B
($69.6\%$ to $69.2\%$) but lowers it in Qwen3.8-27B ($70.5\%$ to $63.3\%$).

\begin{table}[H]
\centering
\small
\setlength{\tabcolsep}{5pt}
\caption{\textbf{Scores on the same candidates in two 27B Qwen models.} Stricter
criterion, $T=1$, no top-$k$ truncation, $16$ raw samples per question; $599$ and
$594$ eligible questions. Each cell gives plurality accuracy (\%) and
pair-weighted AUROC. Masking differences in the text are paired against unmasked
$P(\mathrm{True})$ rescored in the same run, whose plurality accuracy is $33.44\%$
and $27.98\%$.}
\label{tab:qwen27b-rescoring}
\begin{tabular}{lcc}
\toprule
Score & Qwen3.6-27B & Qwen3.8-27B \\
\midrule
Mean log-likelihood & $30.05$ / $0.612$ & $24.75$ / $0.625$ \\
Summed log-likelihood & $29.38$ / $0.604$ & $24.58$ / $0.620$ \\
Chat-template likelihood & $31.22$ / $0.624$ & $25.25$ / $0.643$ \\
PMI & $30.38$ / $0.645$ & $24.24$ / $0.659$ \\
$P(\mathrm{True})$ & $33.39$ / $0.792$ & $28.14$ / $0.773$ \\
Masked $P(\mathrm{True})$ & $30.88$ / $0.612$ & $24.47$ / $0.629$ \\
\bottomrule
\end{tabular}
\end{table}

\begin{table}[H]
\centering
\small
\setlength{\tabcolsep}{5pt}
\caption{\textbf{Paired differences in two 27B Qwen models.} Plurality accuracy in
points and pair-weighted and query-macro AUROC, with paired $95\%$ intervals below. In the last
column, unmasked $P(\mathrm{True})$ is rescored in the same run as masked
$P(\mathrm{True})$.}
\label{tab:qwen27b-differences}
\begin{tabular}{llrrr}
\toprule
Model & Measure & $P(\mathrm{True})-$chat & Masked$\,-\,$chat & Unmasked$\,-\,$masked \\
\midrule
Qwen3.6-27B & Accuracy & $+2.17$ & $-0.33$ & $+2.56$ \\
 & & $[-0.16,4.58]$ & $[-2.84,2.14]$ & $[0.22,5.01]$ \\[2pt]
 & Pair-weighted & $+0.1687$ & $-0.0112$ & $+0.1810$ \\
 & & $[0.1368,0.2009]$ & $[-0.0413,0.0192]$ & $[0.1497,0.2125]$ \\[2pt]
 & Query-macro & $+0.1828$ & $+0.0085$ & $+0.1748$ \\
 & & $[0.1482,0.2178]$ & $[-0.0284,0.0462]$ & $[0.1427,0.2078]$ \\[2pt]
Qwen3.8-27B & Accuracy & $+2.89$ & $-0.79$ & $+3.51$ \\
 & & $[0.57,5.30]$ & $[-3.07,1.46]$ & $[1.11,5.99]$ \\[2pt]
 & Pair-weighted & $+0.1300$ & $-0.0139$ & $+0.1431$ \\
 & & $[0.0935,0.1660]$ & $[-0.0441,0.0168]$ & $[0.1102,0.1748]$ \\[2pt]
 & Query-macro & $+0.1440$ & $-0.0022$ & $+0.1443$ \\
 & & $[0.1037,0.1845]$ & $[-0.0385,0.0346]$ & $[0.1095,0.1782]$ \\
\bottomrule
\end{tabular}
\end{table}

\section{Supporting Ranking Analyses}
\label{sec:resolution}

\subsection{Question structure and readout diagnostics}

In the $512$-entity chat cohort (Appendix~\ref{app:sae}; $2{,}176$ entity--relation rows), $61.9\%$ of entities have both hits and misses under
reference matching, and those entities contain $64.5\%$ of rows. Comparing relations
within an entity fixes entity-level exposure and familiarity but not relation
difficulty: relation identity alone reaches within-entity AUROC $0.9283$, so we
compare candidates only within the same question. Probes cross-fitted on entity-disjoint folds add little beyond relation identity (Table~\ref{tab:resolution}). The output-probability AUROC of $0.4813$ compares greedy chat answers across questions; it differs from the pooled AUROCs of sampled bare-format answers (Appendix~\ref{app:discrimination-tables}) in format, decoding, and population, so the two are not a controlled comparison.
The binding score of a generated answer $\hat o$,
\[
B(s,r,\hat o)=\log P(\hat o\mid s,r)-\log P(\hat o\mid s',r)
-\log P(\hat o\mid s,r')+\log P(\hat o\mid s',r'),
\]
uses no reference answer; its raw AUROC is $0.6697$ and its within-entity AUROC
$0.5915$. A subject-attention-masking analysis attributes $73.8\%$ of within-entity
errors to association with the subject; the remaining subset contains $148$ entities
and $463$ rows, which suggests separating association-driven retrieval from
correctness \citep{cheang2026really}.

\begin{table}[H]
\centering
\small
\setlength{\tabcolsep}{4pt}
\caption{\textbf{Readout diagnostics against relation identity.} Reference-matching
labels. The first, second, and fourth rows are AUROC improvements over relation
identity; the third is a raw AUROC across questions.}
\label{tab:resolution}
\begin{tabular}{llr}
\toprule
Source & Measurement & Result \\
\midrule
Residual state & Last prompt token, $27$ layers & $+0.0015$ \\
Subject attention & $26$ layers, $8$ heads & $-0.0212$ \\
Output probability & Greedy chat answer, raw AUROC & $0.4813$ \\
Binding & Subject-by-relation contrast & $-0.0051$ \\
\bottomrule
\end{tabular}
\end{table}

\paragraph{Entity-level routing.} A policy that makes one answer-or-abstain decision
per entity misroutes at least $\min(h_e,m_e)$ of an entity's $h_e$ hits and $m_e$
misses: $449$ of $2{,}176$ rows ($20.6\%$) in this cohort. An entity-level oracle blocks
$1{,}227$ of $1{,}509$ misses while discarding $167$ of $667$ hits. The bound applies
to policies with one decision per entity, not to relation-aware routing or to the
candidate selection of Section~\ref{sec:selection}.

\subsection{Likelihood ranks answers to the same question}

In the earlier same-question sample (Table~\ref{tab:cohorts}), scored with reference matching, we compare answers to the same question, which holds entity, relation, wording, and every
deterministic pre-generation feature fixed.
Among $245$ retained mixed questions,
there are $5{,}413$ accepted--rejected pairs under reference matching.
Mean log-likelihood achieves pair-weighted AUROC $0.7081$ and
query-macro AUROC $0.7019$. Pooled AUROC on the same answers is $0.6970$, so
likelihood separates answers to the same question at least as well as it
separates answers across questions.

On questions where the greedy answer misses the reference, likelihood still
achieves query-macro AUROC $0.6073$, with $95\%$ interval $[0.5685,0.6475]$
(Appendix~\ref{app:discrimination-tables}, Table~\ref{tab:query-discrimination}).

Ranking and selection can disagree even on a fixed candidate set: if a correct
answer outranks fourteen incorrect answers but loses to a fifteenth, its paired
ranking accuracy is $14/15$, yet argmax returns an incorrect answer. We
therefore report final selection alongside pairwise ranking.

For a deterministic prompt-only score $S(q)$, every within-question comparison
is a tie, giving within-question AUROC $0.5$ by construction. Such a score can still rank
questions: the pre-generation readouts predict $K=16$ coverage across questions
(Section~\ref{sec:know-choose}). Answer conditioning makes
nontrivial within-question ranking possible, and repeated sampling makes it observable
in evaluation; a deployed verifier can still score a single answer.

\subsection{Decoding changes the evaluated response distribution}

In the development sampling cohort, as temperature increases from $0.3$ to
$1.3$, the query-macro AUROC rises by $0.107$ while the pooled AUROC falls by $0.030$
(Appendix~\ref{app:discrimination-tables}, Table~\ref{tab:temperature}).
Accuracy on the retained responses falls at the same time, so better
within-question ranking at higher temperature comes with fewer accepted answers.
Because temperature changes both the candidates and the set of mixed questions,
the trend describes the evaluated response distribution rather than a change in
the scorer.

The pooled statistic has an exact decomposition:
\begin{equation}
A_{\rm pooled}=w A_{\rm within}+(1-w)A_{\rm cross},\qquad
w=\frac{\sum_q n_q^+n_q^-}{(\sum_q n_q^+)(\sum_q n_q^-)}.
\label{eq:auc-decomposition}
\end{equation}
Here $A_{\rm within}$ is pair-weighted, not query-macro, and both components
use the same tie convention. Across the four temperatures of Table~\ref{tab:temperature},
within-question pairs account for $0.06\%$--$0.22\%$ of all comparisons. Pooled AUROC is therefore
dominated by cross-question comparisons; these can carry answer-level evidence, but
pooled AUROC alone cannot isolate within-question ranking.

The same point follows from adding a question-specific offset:
$S'(q,y)=S(q,y)+b(q)$ preserves every within-question ranking and argmax
choice, but can change cross-question rankings and pooled AUROC.

\section{Prompts}
\label{app:prompts}

This appendix reproduces the prompts used in the paper; braces mark fields filled
for each item.

\paragraph{Relation prompts.}
Recall generation and candidate sampling use the relation prompts below as
bare-format prompts, without a chat template; the recall study uses the same prompts for
these ten relations.

\begin{lstlisting}
city_country
  What country contains the city of '{entity}'? The country that contains the city '{entity}' is
player_place_birth
  What is the place of birth for the player '{entity}'? The city of birth for the player '{entity}' is
player_teams_list
  What team (name at least one) signed the player '{entity}'? The team that signed the player '{entity}' is
movie_cast
  What is the name of an actor starring in the movie '{entity}'? The actor starring in the movie '{entity}' is
movie_directors
  Who is the director of the movie '{entity}'? The director of the movie '{entity}' is
movie_screenwriters
  Who is the screenwriter of the movie '{entity}'? The screenwriter of the movie '{entity}' is
song_album
  What is the album of the song '{entity}'? The album of the song '{entity}' is
song_performers
  Who is the performer of the song '{entity}'? The performer of the song '{entity}' is
song_genres
  What is the genre of the song '{entity}'? The genre label that best describes the song '{entity}' is
song_publication_year
  What is the publication year of the song '{entity}'? The publication year for the song '{entity}' is
\end{lstlisting}

\paragraph{Readout prompts.}
The Gemma and Llama readouts are read at the final entity token of the entity-only
prompt below; the Qwen3 feature is read at the final entity mention of the relation
prompt.

\begin{lstlisting}
The {entity_type} '{entity}'
\end{lstlisting}

\paragraph{Verification prompts.}
The prompt below fills the user turn of the model's chat template, followed by the
generation prompt; the candidate answer is whitespace-collapsed and truncated to
$300$ characters, and Qwen's thinking mode is disabled. The question is the relation prompt up to
its question mark. $P(\mathrm{True})$ normalizes the first-response probabilities
of the affirmative and negative tokens (\emph{Yes}/\emph{No}, or \emph{True}/\emph{False}).

\begin{lstlisting}
Question: {question}
Proposed answer: {answer}

{instruction}
\end{lstlisting}

The original instruction and the three wordings of
Appendix~\ref{app:prompt-robustness} are:

\begin{lstlisting}
original:    Is the proposed answer factually correct? Reply Yes or No.
short:       Is this answer correct? Reply Yes or No.
true/false:  Is the proposed answer factually correct? Reply True or False.
no/yes:      Is the proposed answer factually correct? Reply No or Yes.
\end{lstlisting}

Each label is the single token that the model's tokenizer assigns to the word
without a leading space, read at the first response position; no case, whitespace,
punctuation, or markdown variants are merged. Table~\ref{tab:label-tokens} lists
the token IDs recorded by the scoring runs.

\begin{table}[H]
\centering
\small
\setlength{\tabcolsep}{6pt}
\caption{\textbf{Label token IDs.} \emph{True} and \emph{False} are used only by the
true/false wording, which is evaluated for Gemma and Llama.}
\label{tab:label-tokens}
\begin{tabular}{lrrrr}
\toprule
Model & \emph{Yes} & \emph{No} & \emph{True} & \emph{False} \\
\midrule
Gemma~2 2B-IT & $3553$ & $1294$ & $5036$ & $8393$ \\
Llama~3.1 8B-Instruct & $9642$ & $2822$ & $2575$ & $4139$ \\
Qwen3-8B & $9454$ & $2753$ & --- & --- \\
\bottomrule
\end{tabular}
\end{table}

\paragraph{Rescoring prompt.}
Chat-template likelihood (Appendix~\ref{app:rescoring}) scores the candidate as the
assistant reply to the following user message, rendered with the model's chat
template. PMI and masked $P(\mathrm{True})$ replace the entity in the
question with \emph{this \{type\}}.

\begin{lstlisting}
Return only the requested answer, without alternatives or explanation. Question: {question}
\end{lstlisting}

\paragraph{Semantic judge prompts.}
Pass~0 (Appendix~\ref{app:semantic}) receives only the question and its reference
answers and uses the following prompt.

\begin{lstlisting}
You are auditing the validity of a factual question and its reference answer set before any candidate responses are evaluated.

Your task is NOT to answer the question from scratch and NOT to judge any model response.
You must determine whether the supplied Question and Reference Answer(s) form a valid, sufficiently clear evaluation target.

You will see only:
- the factual Question,
- the Reference Answer(s).

You will NOT see candidate answers, model scores, selector identities, or previous automatic labels.

Use the following rules.

1. VALID
Return VALID when:
- the question has a clear factual interpretation;
- the supplied reference answer(s) are factually compatible with that question;
- the reference set is adequate for evaluating ordinary candidate answers under the stated relation semantics.

Aliases, spelling variants, abbreviations, and equivalent names do not make a reference invalid.

2. INVALID_OR_AMBIGUOUS
Return INVALID_OR_AMBIGUOUS when any of the following holds.

- The supplied reference answer is clearly wrong for the question, or clearly corresponds to a different entity, relation, work, date, or factual target. Do not use this label merely because another answer may also be valid.

- The question itself does not specify a unique or sufficiently well-defined factual target for the intended relation. Examples include:
  - missing temporal scope when the answer materially depends on time;
  - unresolved entity ambiguity;
  - wording that could reasonably refer to different relations;
  - a question whose singular/plural requirement is unclear in a way that changes correctness.

- The relation is genuinely multi-valued, AND you are highly confident that the supplied reference set omits one or more other clearly valid answers that an otherwise correct candidate could reasonably give. Do not require exhaustive reference sets for single-answer questions. Do not flag a multi-valued relation merely because additional obscure or historical values might exist.

Relation semantics:
- If the question asks for one valid value from a multi-valued relation, one valid answer is sufficient.
- If the question explicitly asks for all values, the reference set must be exhaustive enough to support that requirement.
- Publication-year questions require the intended publication year for the named work.
- For names, aliases and equivalent forms refer to the same factual answer.

Be conservative.
If the supplied reference is a standard, plausible answer and there is no clear defect that would make candidate evaluation unreliable, return VALID.

Do not repair, expand, or rewrite the reference answer set.

Return exactly one JSON object:
{"reference_status":"VALID"}
or
{"reference_status":"INVALID_OR_AMBIGUOUS"}

Question:
{QUESTION}

Reference Answer(s):
{REFERENCE_ANSWERS}
\end{lstlisting}

Pass~1 uses the following prompt.

\begin{lstlisting}
You are a blinded evaluator of factual question-answering responses.

Your task is to determine whether the Candidate Response correctly answers the Question, using the Reference Answer(s) as the accepted factual answer set.

You must judge the semantic meaning of the complete Candidate Response, not whether it merely contains a reference string.

Rules:

1. A response is CORRECT if it clearly gives a factually correct answer to the question and is semantically consistent with at least one accepted reference answer.

2. Aliases, abbreviations, spelling variants, paraphrases, and equivalent formulations count as correct when they clearly refer to the same answer.

3. A response that lists multiple alternatives without resolving which one is the answer is INCORRECT.
   Example: "either Lebanon or Syria" is incorrect for a question whose answer is Lebanon.

4. A response may mention multiple alternatives and still be CORRECT if it clearly and unambiguously identifies the correct one as its final answer.
   Example: "It could be A or B, but the correct answer is A" is correct when A is correct.

5. Hedging alone does not make an answer incorrect.
   "Probably Lebanon" is correct if Lebanon is the correct answer and no competing answer is asserted.

6. A refusal, non-answer, irrelevant response, or response that never states an answer is INCORRECT.

7. If the response contradicts the correct answer, it is INCORRECT even if the correct reference string also appears elsewhere in the response.

8. For a relation with multiple valid answers:
   - Any one accepted value is sufficient unless the question explicitly asks for all values, a complete list, or otherwise requires exhaustive enumeration.
   - If several values are asserted as answers, all asserted values must be valid. A mixture of correct and incorrect asserted values is INCORRECT.
   - A list of several possibilities with no resolved answer is INCORRECT.

9. For single-valued questions, giving multiple incompatible answers without resolving to the correct one is INCORRECT.

10. For publication-year questions, the stated year must match the accepted year. Equivalent textual forms are allowed, but an unresolved range or multiple competing years is INCORRECT.

Judge only the factual correctness of the answer. Do not judge style, fluency, confidence, verbosity, or whether the response resembles the reference wording.

Do not infer anything from model identity or selection method. You will not be given model scores or selector identities.

Return exactly one JSON object:
{"label":"CORRECT"}
or
{"label":"INCORRECT"}

Question:
{QUESTION}

Reference Answer(s):
{REFERENCE_ANSWERS}

Candidate Response:
{EXACT_CANDIDATE_TEXT}
\end{lstlisting}

\end{document}